\documentclass[runningheads]{llncs}
\usepackage{graphicx}
\usepackage{amsmath}
\usepackage{comment}
\usepackage{subcaption}
\usepackage{float}
\begin{document}
\title{Opening the Black Box: Analyzing Attention Weights and Hidden States in Pre-trained Language Models for Non-language Tasks} 
\titlerunning{Analyzing Attention for Non-language Tasks}
%

\author{Mohamad Ballout \and Ulf Krumnack \and Gunther Heidemann \and Kai-Uwe Kühnberger}

\institute{Institute of Cognitive Science, University of Osnabrück, Osnabrück, Germany}
\authorrunning{Mohamad Ballout et al.}
%
%

\maketitle              
\begin{abstract}

Investigating deep learning language models has always been a significant research area due to the ``black box" nature of most advanced models. With the recent advancements in pre-trained language models based on transformers and their increasing integration into daily life, addressing this issue has become more pressing. In order to achieve an explainable AI model, it is essential to comprehend the procedural steps involved and compare them with human thought processes. Thus, in this paper, we use simple, well-understood non-language tasks to explore these models' inner workings. Specifically, we apply a pre-trained language model to constrained arithmetic problems with hierarchical structure, to analyze their attention weight scores and hidden states. The investigation reveals promising results, with the model addressing hierarchical problems in a moderately structured manner, similar to human problem-solving strategies. Additionally, by inspecting the attention weights layer by layer, we uncover an unconventional finding that layer 10, rather than the model's final layer, is the optimal layer to unfreeze for the least parameter-intensive approach to fine-tune the model. We support these findings with entropy analysis and token embeddings similarity analysis. The attention analysis allows us to hypothesize that the model can generalize to longer sequences in ListOps dataset, a conclusion later confirmed through testing on sequences longer than those in the training set. Lastly, by utilizing a straightforward task in which the model predicts the winner of a Tic Tac Toe game, we identify limitations in attention analysis, particularly its inability to capture 2D patterns. 

\keywords{Pre-trained language model  \and Transformers \and XAI \and Attention analysis \and BERT}
\end{abstract}
\section{Introduction}

Pre-trained Language models, based on transformer architecture \cite{transformer-ref-1}, have surpassed previous benchmarks in the majority of language-related tasks, demonstrating their superiority over earlier sequence networks like RNNs. Models such as BERT \cite{BERT-ref-4}, T5 \cite{T5-ref-5}, BART \cite{BART-ref-6}, and GPTs \cite{GPT2-ref-7} are now extensively employed in various language tasks and have expanded into other domains, including vision \cite{vit-ref-2} and audio processing \cite{audio-ref-3}. One contributing factor to the success of transformers is their attention mechanism, a feature that essentially allows the model to focus on and assign different degrees of importance to various parts of an input sequence when generating the output. This mechanism not only brings inherent interpretability but also allows researchers to visualize and comprehend the relationships between different components of the transformers, providing insights into how the model processes and weighs various features during its decision-making process. However, as these models continue to grow in scale, featuring billions of parameters, the longstanding challenge of deep learning's explainability resurfaces. Can we truly comprehend the reasoning behind a model's predictions?

Numerous studies have been conducted to reveal and interpret the network output on language tasks. Earlier research focused on attention analysis within the linguistic context \cite{X-ref-8,Dark-ref-9,inter-ref-10}. However, with the advancements in these models, such as GPT-4's \cite{gpt4-ref} demonstrated ability to tackle various problem types including arithmetic and logical reasoning, we explore the interpretability of pre-trained language models in non-language tasks. Many prior studies have encountered challenges when attempting to examine explainability in language data due to its inherent complexity, often yielding inconclusive results. Some previous work went further, stating that attention analysis could not be perceived as an explanation for the model's decisions \cite{not-X-ref-11}. To our knowledge, there is a lack of literature exploring the analysis of pre-trained language models in non-linguistic tasks. Therefore, in this study, we fine-tune BERT on non-linguistic tasks, such as arithmetic datasets like ListOps \cite{LRA-ref-23}, and to predict the winner of a basic Tic Tac Toe game, and examine the outcomes of attention heads and token representations. Our findings reveal interesting results that were not apparent when examining attention in language contexts. For example, in numerous instances, it was possible to deduce the model's predicted response solely by examining the attention heat maps. Furthermore, by analyzing the attention mechanism, we hypothesize that BERT's approach to solving an hierarchical arithmetic tasks closely resembles the process a human would follow.

Our examination centered on BERT, a pre-trained language model that is extensively employed in the literature. BERT uses a bidirectional context encoder for learning textual representations. We aim to investigate BERT's attention and representation in tasks unrelated to language to better understand the workings of pre-trained language models. Our objective is to enhance the models' interpretability, applicability, and reliability across various domains. This analysis as, shown later in the paper, lead to more efficient fine-tuning, better understanding of the model's decision making and its limitations. Upon publication, the code, model, and datasets used in our experiments will be made available as open source resources.\footnote{https://github.com/BalloutAI/Attention-Analysis.git}

The main contributions of this paper are:
\begin{itemize}

\item
Conducting a comprehensive interpretability assessment of the BERT model fine-tuned on three non-linguistic tasks, employing four visualization methods: token-to-token analysis, attention heatmap analysis, token embeddings similarity analysis, and entropy analysis. In addition, by performing a layer-by-layer examination using these methods, we are able to make inferences about decision-making processes in pre-trained language models.

\item
In contrast to attention analysis on language data, where drawing conclusions from visual representations can be challenging, we find that the various techniques used in this study provided clear evidence of the attention models' effectiveness. We observe clear examples demonstrating the effectiveness of attention models, as we could discern the model's predictions from the heatmaps or token embeddings.

 \item
The explainability analysis enabled us to formulate hypotheses about the model's generalization capabilities and fine-tuning techniques, which were later confirmed through testing.  

\end{itemize}

\section{Related Work}

There is extensive research in the literature focused on analyzing the attention mechanisms within transformer models, with many researchers attempting to explain and examine the output of these models. For example, \cite{Exbert-ref-16}  introduced a visualization tool designed to investigate the attention weights and contextual representations of transformer-based models. This tool represents the attention weights of each token relative to other tokens using curved lines, allowing users to explore attention within input sequences. A similar tool was also developed by \cite{bertviz-ref-17} for visualizing self-attention. In this work, we refer to this type of analysis as token-to-token analysis to differentiate it from heat map analysis.

Many previous studies have concentrated on identifying the linguistic skills acquired by pre-trained language models. These studies typically analyze attention weights to interpret such linguistic abilities by carefully selecting specific inputs and examining the corresponding outputs. For instance, \cite{Asses-ref-18} found that BERT consistently assigned higher scores to correct verb forms when fed both correct and incorrect subject-verb compositions, suggesting that BERT possesses syntactic capabilities. Similarly, the study in \cite{structure-ref-19} proposed using probing tasks, which are diagnostic tasks specifically designed to examine the representation of particular types of linguistic information within a model. These tasks are employed to explore the unique properties and characteristics of linguistic details, including phrase-level information, syntactic structures, and semantic relationships, as encoded by different layers within the BERT model.

\cite{Look-ref-20} conducted a more in-depth investigation into understanding what aspects BERT focuses on, discovering common behaviors across attention heads. While no attention head excelled at multiple syntactic relations, specific heads were found to correspond well with particular semantic and syntactic relations like finding direct objects of verbs, determiners of nouns, etc. In \cite{Reveal-ref-21},  the authors demonstrated BERT's capacity to capture various types of linguistic information. Furthermore, they utilized their analysis to improve BERT's performance on language tasks, demonstrating absolute performance increases of up to 3.2\% on specific tasks, such as Recognizing Textual Entailment (RTE).

Conversely, some studies in the literature express skepticism about the impact of attention weights on the system's output, as seen in \cite{notX-ref-12,isNX-ref-23}. Both studies observed that higher attention weights do not significantly influence model predictions. In fact, \cite{notX-ref-12} concluded that attention modules do not offer meaningful explanations for the model's output. However, \cite{isX-ref-14} challenged the tests conducted in  \cite{notX-ref-12} and proposed alternative tests to determine if attention can be used for explanation. The authors claim that their proposed tests disprove the assertions made in  \cite{notX-ref-12}.

Another area of research closely related to this work involves utilizing pre-trained language models for non-linguistic tasks \cite{bert-ref-134,MathBert-ref-144,universal-ref-15}. For instance, \cite{MathBert-ref-144} proposed a novel pre-trained model called MathBERT, trained with both mathematical formulas and their contexts. The empirical findings show that MathBERT substantially surpasses the performance of current techniques on all evaluated math-related tasks. In addition, \cite{universal-ref-15} demonstrated that pre-trained language models can serve as universal computational engines to solve tasks across various domains, requiring only 0.1\% of GPT-2's parameters.
This work aims to explain the remarkable success of pre-trained language models in these non-language tasks.  
      
In light of this discussion and considering that linguistic probing tasks have been thoroughly explored in the literature, we build upon previous work conducted in the language context, analyzing pre-trained language transformers on arithmetic tasks. Our analysis includes both local and global attention weight analysis, which means we examine individual tokens and their target attention, as well as the attention heatmap for the entire sequence.

We demonstrate that this type of task clearly highlights the importance of attention and, in many instances, reveal decision-making processes in a pre-trained language model like BERT. We observed that the model often does not attend to tokens that do not influence the results, whereas attention is focused on tokens that are crucial for the output. Additionally, we investigated token embeddings and their similarity to each other across layers. We found that embedding similarities align with attention analysis, as the decisive tokens—those that determine the model's output—exhibit higher similarities than tokens deemed less important.

We also discovered that the last layers, especially and surprisingly the 10th layer rather than the last one, in the encoder transformer encodes the most apparent attention weights, allowing us to determine the answer by simply examining the attention maps. Based on these findings, we compared fine-tuning of BERT in two settings: by freezing all parameters except those in layers 10 and 12. We discovered that unfreezing the parameters in layer 10 outperforms unfreezing them in layer 12. Overall, this method reduced the number of parameters in the base version of BERT, which originally had approximately 109 million parameters, to just 7 million. Despite this reduction, we achieved competitive results with only a minor decrease in accuracy on non-linguistic tasks in both of the mentioned fine-tuning settings.

\section{Model and Datasets}

In this section, we briefly discuss the model used which is BERT, the datasets used to test the model, and the fine-tuning process.

\subsection{Model-BERT}

We chose to analyze BERT, or Bidirectional Encoder Representations from Transformers, on non-language tasks because it is one of the most renowned and early pre-trained language models introduced by \cite{BERT-ref-4}. It is based on the transformer architecture by \cite{transformer-ref-1}, utilizing the encoder component exclusively. BERT's bidirectional design facilitates the simultaneous consideration of both, the preceding and following context in a given text, resulting in a more nuanced and comprehensive representation of linguistic information. This model is extensively employed for fine-tuning a wide range of NLP tasks, including but not limited to question answering \cite{qa-ref-25}, sentiment analysis \cite{sent-ref-26}, and named entity recognition \cite{BERT-ref-4}.

\subsection{Datasets} 

We choose to evaluate BERT on three non language datasets including two versions of the ListOps and a Tic Tac Toe game winner prediction. The reasoning behind selecting these tasks is that we require challenges for which we possess a clear understanding of the solution process, as opposed to language and vision tasks, where the human brain's processing mechanisms remain less comprehensible. The three non-language datasets are: 
\begin{itemize}
\item

\textbf{ListOps}

The Long ListOps dataset is a classification task initially developed by \cite{listops-ref-22} and later extended by \cite{LRA-ref-23} to create longer sequences in order to test transformers in long-range scenarios. Designed to test a model's capability to handle hierarchical data in a longer setup, the dataset comprises mathematical operations on natural numbers such as Modular Addition, Minimum, Maximum, and Median. A short example of a sequence is:

$${[\text{SUM } \ [\text{MIN} \ 0 \ 1 \ [\text{MED} \ 5 \ 2 \ 1 \ ] \ 4 \ ] \ 1 \ 2 \
 [\text{MAX} \ 1 \ 5 \ 4 \ ] \ ]} $$ 

To solve this sequence, the model must process it hierarchically, as some sub-sequences within the longer sequence need to be resolved first, as illustrated in figure \ref{fig1}. Additionally, in order to obtain the correct answer, the model must attend to every token in long scenarios. Previous studies have shown that transformers have difficulties in processing this type of data, as the model must remember the order of every token to accurately predict the answer. The dataset features a 10-class classification, where the classes are the integers from 0 to 9. For simplicity, we limit the range of the number of tokens per sequence to 200-400. The model is trained on 98k sequences and tested on 2k samples adopting this approach from the original work by \cite{LRA-ref-23}. BERT scores an accuracy of 61.6\% on this dataset.

\begin{figure}
\centering
\includegraphics[scale=0.6]{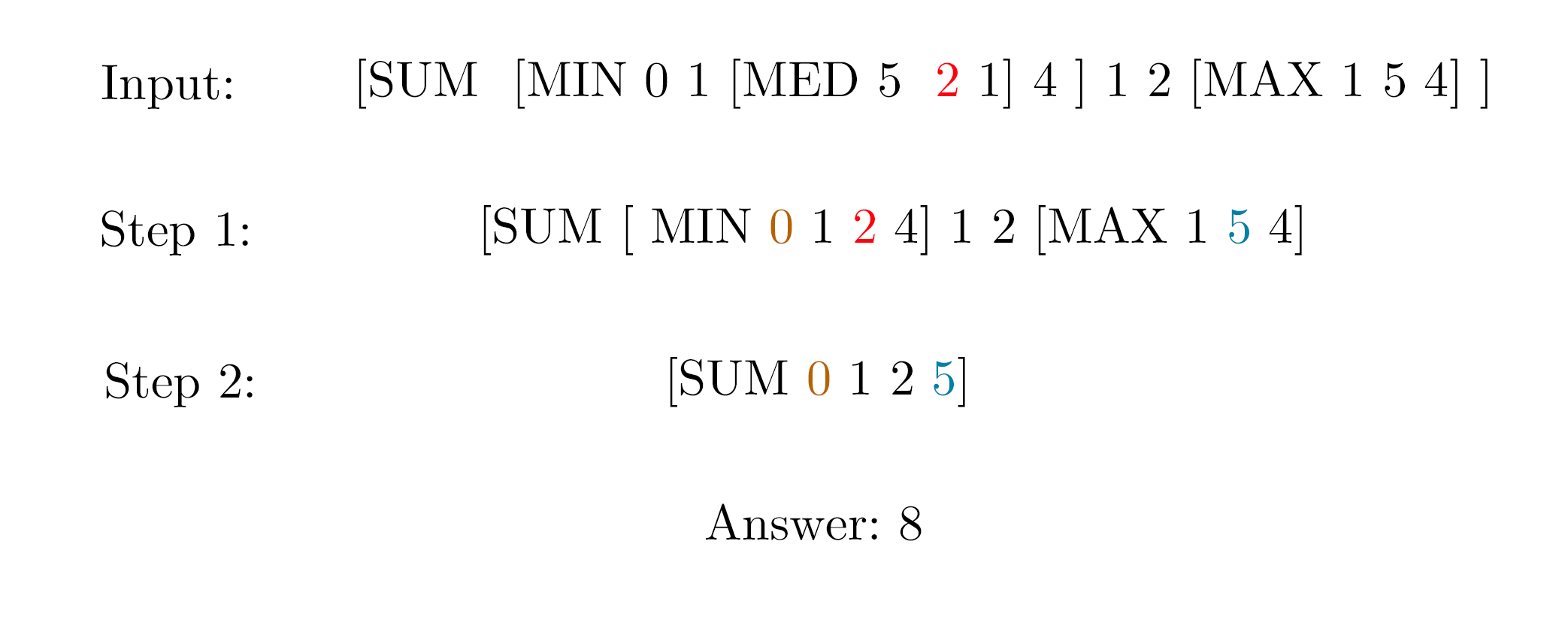}
\caption{ Figure illustrates the standard process of solving a sequence in ListOps.} 
\label{fig1}
\end{figure}
\item

\textbf{Modified ListOps}

This task is similar to the original ListOps \cite{LRA-ref-23}, but we modified it to create an easier task. The primary reason for this modification is that we observed the model's inability to solve the modular addition, an intriguing problem worth investigating in the future. To enable a clearer  attention analysis, where the model can solve all operators, we replaced the SUM and MED operators with simpler ones, including FIRST and LAST. The First and Last operators require the model to simply choose the first and last elements in the sequence or sub-sequence, respectively. A short example of a sequence would be:

 $${[\text{LAST } \ 2 \ [\text{MIN} \ 0 \ 5 \ 4 \ ] \ 3 \ [\text{FIRST} \ 5 \ 1 \ ] \ 4 \ 9 \ [\text{MAX} \ 1 \ 5 ]\ ]} $$ 

In this example, the answer is the last element in the sequence, which is 5. It is important to note that the actual training and testing samples have lengths ranging from 200 to 400, making the problem more complex. The model scores an accuracy of 95.1\% on this dataset.

\item

\textbf{Tic Tac Toe}

Finally, the model is fine-tuned to determine the winner of a Tic Tac Toe game. This is a binary classification problem, and the sequence input consists of a flattened 1D representation of the game. An example input is as follows:

$$\begin{tabular}[c]{c|c|c}\mbox{\ -\ }&\mbox{\ x\ }&\mbox{\ x\ }\\\hline x&x&o\\\hline o&o&o\end{tabular}\quad\to\quad \ - \ x \ x \ |  \ x \ x \ o \ | \ o \ o \ o \ | $$ 

The symbol `` - " to an unfilled space whereas the vertical bar is the delimiter of each dimension. In the example shown above the correct output is \textit{o}, since the player filled the third row completly. This is an easy task where the model scores 100\%.  

\end{itemize}

\subsection{Fine-tuning BERT}

\begin{figure}[htbp]
    \centering
    \includegraphics[scale=0.48]{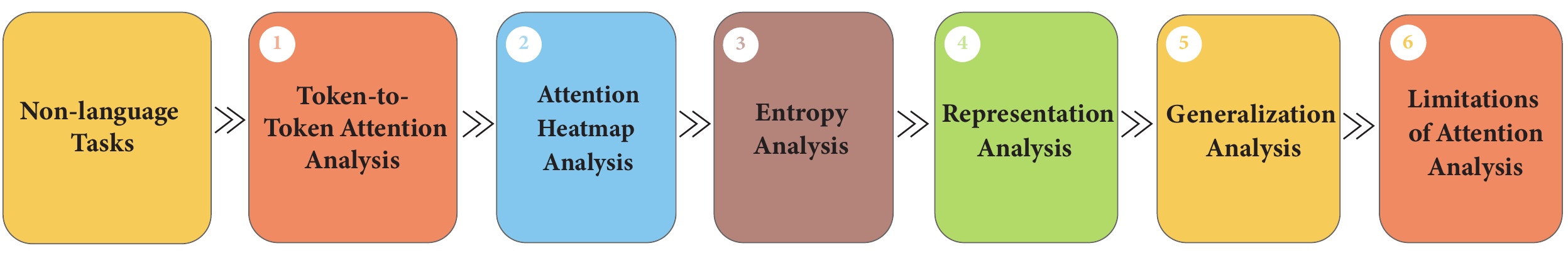}
    \caption{Figure shows the type of analysis done on non-language tasks.}
    \label{fig01}
\end{figure}

In our experiments, we fine-tuned the BERT model on all of the tasks above using the Hugging Face \cite{Hugg-ref-24} implementation of the BERT-base model. The model consists of 12 layers (i.e., 12 transformer blocks) and uses 12 attention heads in each of those layers. We fine-tuned BERT with learning rate of 2e-5 and the remaining hyper-parameters were adopted from \cite{arithm-ref-22}. We fine-tune all of the model's parameters, totaling approximately 109 million, except in certain cases where we specify that we freeze all parameters except for a particular layer. In such cases, we freeze all parameters with the exception of the specified layer and the norm layer in all layers. This training strategy was adopted from \cite{universal-ref-15}.  We  follow the data splitting approach for training and validation from \cite{LRA-ref-23}, which trains on 98\% of the data and validates on the remaining 2\%. The various analyses we have performed are summarized in figure \ref{fig01}.

\section{Analysis and Results}

We visualize the attention mechanisms in BERT using two methods: first, by employing the ExBert tool provided by \cite{Exbert-ref-16}, and second, by generating heatmaps of the attention score matrix. Both approaches offer valuable insights into the relationships and dependencies among input tokens. ExBERT provides an attention view option that enables us to examine each token's attention and its targets, while the attention heatmaps offer a comprehensive snapshot of the attention patterns in a single image. Both methods utilize attention weights, denoted by the matrix $\boldsymbol{\alpha}$, which are computed using the scaled dot-product attention formula:

\begin{equation}
\alpha_{ij} = \frac{\exp(\mathbf{q}_i^T \mathbf{k}j)}{\sum_{l=1}^{n} \exp(\mathbf{q}_i^T \mathbf{k}_l)}
\end{equation}

Here, $\alpha_{ij}$ denotes the attention weight between the $i$-th query and the $j$-th key, signifying the importance assigned by the model to each input token when generating the output token. By visualizing the attention weights matrix $\boldsymbol{\alpha}$, we can visually interpret the model's focus and the underlying structure of the token relationships. Each element in the matrix corresponds to the attention weight between the $i$-th and $j$-th tokens, effectively capturing the dependencies among them. 

Another important aspect to consider is that BERT models introduce special tokens (e.g.,``[CLS]", ``[SEP]") for downstream classification or generation tasks. The ``[CLS]" token (short for ``classification") is added at the beginning of the input sequence and is used to aggregate information from the entire input sequence for tasks that require a single output, such as classification or sentiment analysis. The ``[SEP]" token (short for ``separator") is used to separate different parts of the input sequence, such as individual sentences in a sentence-pair classification task or the context and the question in a question-answering task. These tokens often receive substantial attention and function as a null operation, as explained by \cite{Look-ref-20}. We followed the suggestion by \cite{Exbert-ref-16} to hide these special tokens of the model and re-normalize based on the other attentions to provide easier visualization of subtle attention patterns.

\subsection{Token-to-Token Analysis} 

We start our analysis by exploring the token-to-token mappings in a fine-tuned BERT model on ListOps, utilizing the ExBERT tool. The visualizations reveal interesting findings. In the network's initial layers, each operator (e.g. ``MAX", ``MIN", etc.) attends to its corresponding numbers, neighbors, and brackets, while in the last layers, all tokens focus on the first operator and/or the answer.
It is crucial to recognize that, in the ListOps dataset, the sequence's initial operator is responsible for predicting the answer and should be resolved last. We propose that this observation indicates BERT's ability to comprehend the problem's hierarchical nature and resolve it in a manner similar to the one depicted earlier in figure \ref{fig1}.

Given the short sequence for the sake of the visualization:

$${[\text{MAX } \ 2 \ 3 \  [\text{MIN} \ 1 \ 5 \ 6 \ 1 \ 2 ] \ 1 \ [\text{FIRST} \ 1 \ 4 \ 2 \ ] \ 8 \ ] }$$ 

In the initial layers, ranging from 1 to 4, it is apparent that the tokens attend to their neighbors without recognizing the boundaries established by brackets and sub-sequences. This suggests that the tokens lack a well-defined objective. We hypothesize that, in these early layers, the tokens attempt to discern their positions within the sequences.

\begin{figure}[htbp]
    \centering
    \includegraphics[scale=0.35]{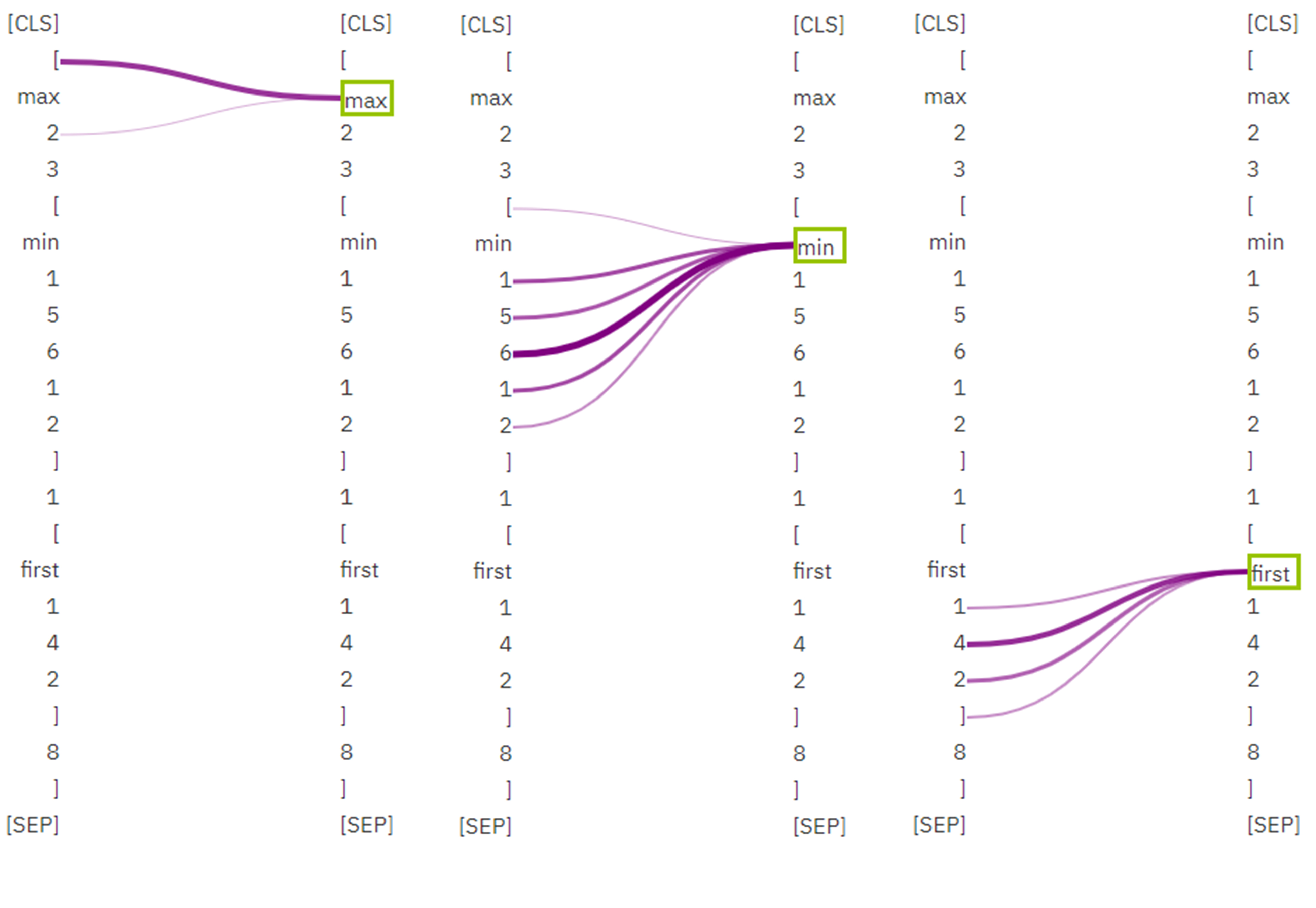}
    \vspace{-0.5cm}
    \caption{Figure shows that operators inside each sub-sequence are attended by their tokens in layer 6.}
    \label{fig3}
\end{figure}

In layers 5 and 6, it becomes evident that the model has identified the boundaries of each sub-sequence. Consequently, in these layers, for the provided example, we observe that tokens attend to their respective operators. For instance, the ``MAX" operator is attended by token 2, while the ``MIN" operator is attended by tokens 1, 5, 6, 1, and 2, and the ``FIRST" operator is attended by tokens 1, 4, and 2. It is important to note that this is not always flawless, as evidenced by the maximum operator not attending to token ``3" in this case. Furthermore, in some instances, the opening and closing brackets also attend to their corresponding operators. Figure \ref{fig3} illustrates the distribution of attention weights in layer 6, given the same input sequence as above. The green rectangles surrounding the tokens signify that we are exclusively displaying the attentions correlated with these specific tokens. 

\begin{figure}[htbp]
\includegraphics[scale=0.35]{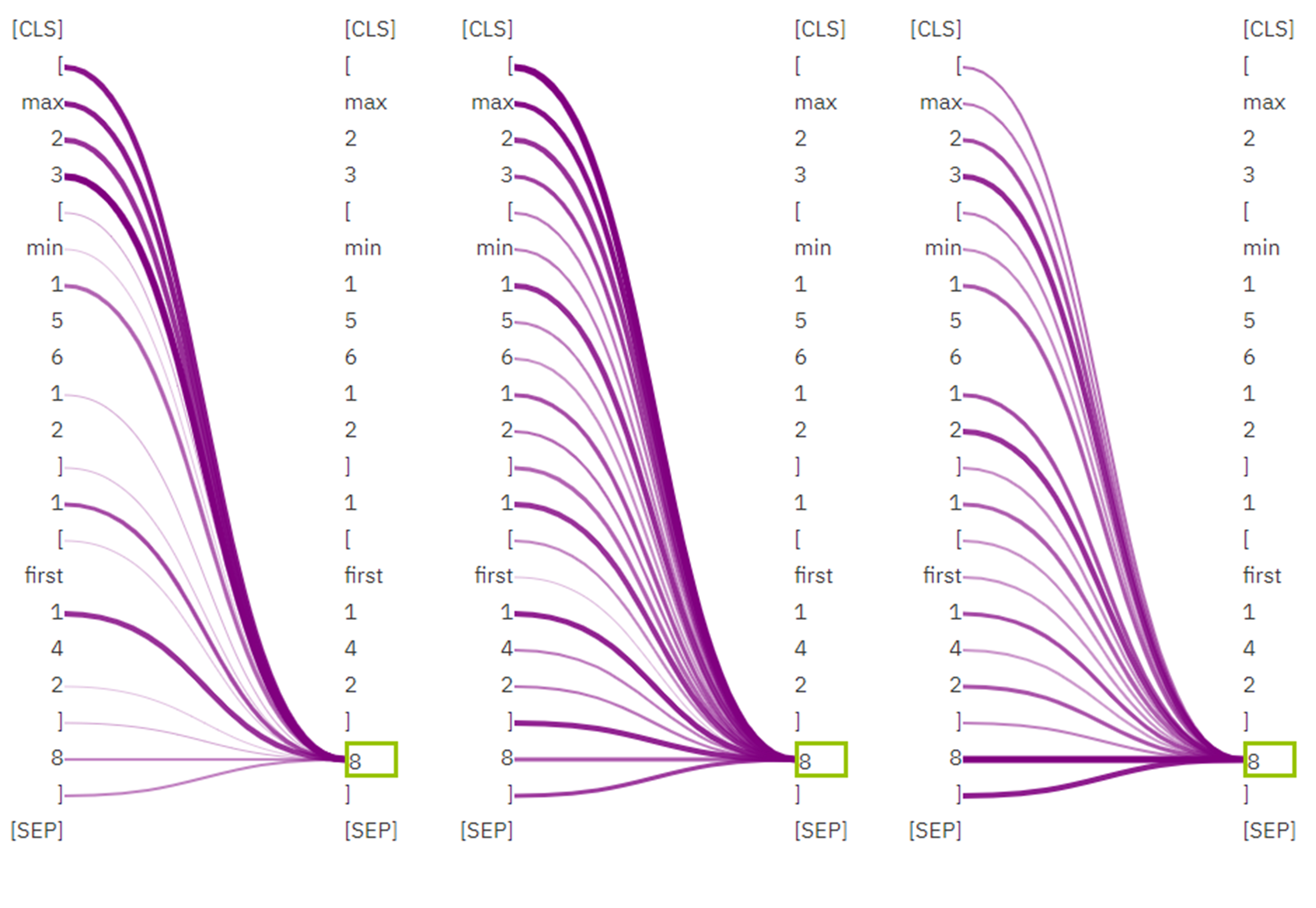}
\caption{ Figure shows that most of the tokens attend to the correct answer in layers 10, 11, and 12 from left to right  } 
\label{fig4}
\end{figure}

From layers 7 to 12, it is strikingly apparent that the system's focus shifts towards determining the correct answer. As a result, we observe that the majority of attention is directed to the correct answer in most of these layers. The most attended token is the correct answer, and it is therefore obvious that one could predict the system's output simply by examining these attention patterns. The attention of tokens towards the correct answer, which is 8 in the given example, is illustrated in layers 10, 11, and 12 in figure \ref{fig4}.

In figure \ref{fig4}, we only illustrate the attention given to the actual answer tokens, but other tokens also receive attention. Upon closer inspection of layers 10, 11, and 12 within the model, using various examples, we uncover some interesting features. It appears that the tokens receiving the most attention tend to be the answers for each sub-sequence. This simplification step bears resemblance to step 2 in the process shown in figure~\ref{fig1} of how a person might approach solving the problem. This particular behavior is prominently observed in layer 10. For example, consider the following sequence:

$${[\text{FIRST } \ 2 \ 3 \  [\text{Max} \ 1 \ 5 \ 6 \ 1 \ 2 ] \ 0 \ [\text{MIN} \ 1 \ 0 \ 2 \ ]\ ] }$$ 

The simplified problem after solving the sub-sequences would become:  

$${[\text{FIRST } \ 2 \ 3 \  6 \ 0 \  0 \ ] }$$ 

\begin{figure}[htbp]
    \centering
    \begin{subfigure}{0.45\textwidth}
        \centering
        \includegraphics[scale=0.5]{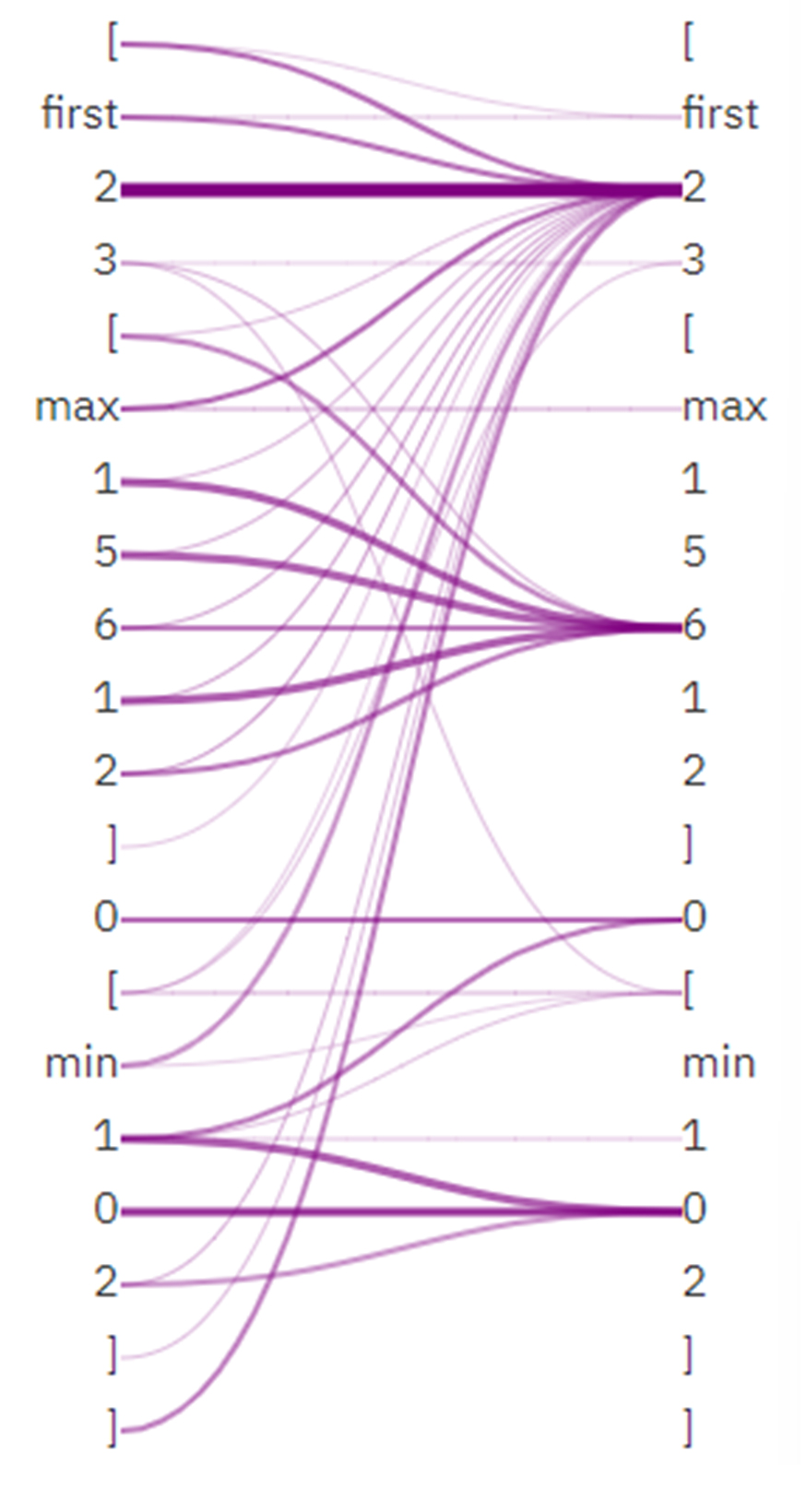}
        \caption{Most attended tokens are the answers of sub-sequences and the answer of the whole sequence ``2"}
        \label{fig5}
    \end{subfigure}
    \hfill
    \begin{subfigure}{0.45\textwidth}
        \centering
        \includegraphics[scale=0.5]{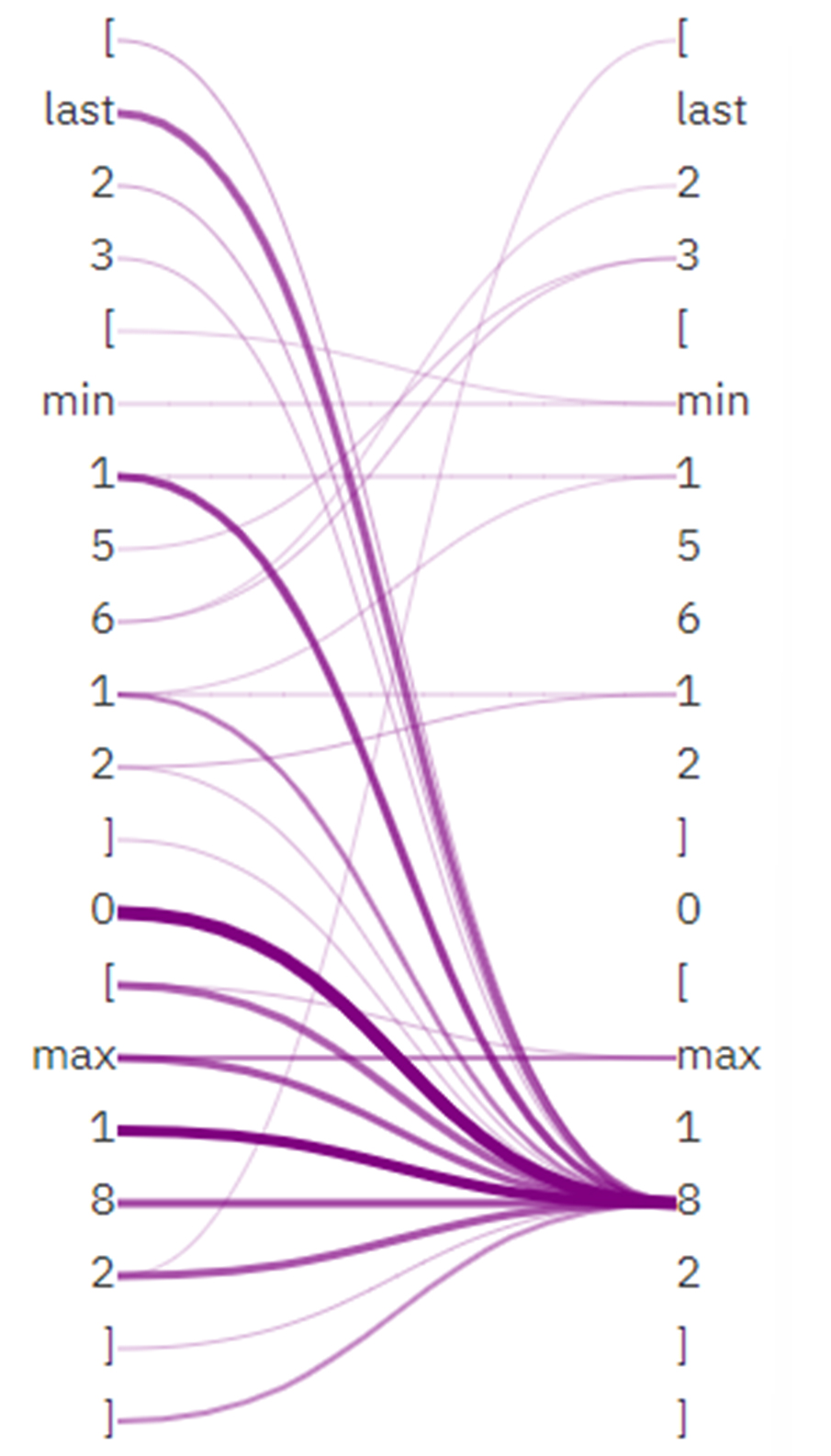}
        \caption{Most attended tokens are the answers of sub-sequences and the answer of the whole sequence ``8"}
        \label{fig6}
    \end{subfigure}
    \label{fig:subfigures}
    \caption{}
\end{figure}

The tokens of the simplified version match the most attended tokens in layer 10, with the token ``2" receiving the highest attention. This token is the correct answer, as demonstrated in figure \ref{fig5}.

A more challenging instance occurs when the answer is not as apparent as in the previous example, as it is embedded within a sub-sequence. Nevertheless, we observe a similar pattern in which the model nearly attends to the simplified sequence version, with one token missing only. The example is presented as follows:

$${[\text{LAST } \ 2 \ 3 \  [\text{MIN} \ 1 \ 5 \ 6 \ 1 \ 2 ] \ 0 \ [\text{MAX} \ 1 \ 8 \ 2 \ ]\ ] }$$ 

The simplified problem after solving the sub-sequences would become  

$${[\text{LAST } \ 2 \ 3 \  1 \ 0 \  8 \ ] }$$

As depicted in figure \ref{fig6}, the most attended tokens are nearly identical to the simplified version, with the exception of 0. Once again, we observe that the predominantly attended token is ``8" which represents the correct answer. It is crucial to acknowledge that this is not always the case, but we have noticed that it is a common pattern in layer 10. Interestingly, we discovered that this distinct pattern is not evident in layer 12, where we initially anticipated clearer observations. Although the tokens in layer 12 still attend to the correct answer, we detect more noise in the attention given to tokens that are not crucial for determining the answer.

In light of this unanticipated finding, we experimented with a novel fine-tuning technique as an alternative to the traditional method. Conventionally, the least parameter-intensive approach to fine-tuning a model involves freezing all model parameters except for the last layer. However, given our observation that the attention in layer 10 is clearer and more easily explained than in layer 12, we compared the model performance under two conditions: freezing all parameters except layer 10 and freezing all parameters except layer 12. The results presented in table \ref{tab1} are both surprising and unconventional. Although both options yield lower scores than the fully fine-tuned model, the version fine-tuned with layer 10 outperforms the one with layer 12 by 5.3\%  on the modified ListOps dataset, with respective scores of 86.1\% and 80.8\%. It also outscores on the original ListOps dataset with respective scores of 54.5\% and 52.8\%. 

\begin{table}[htbp]
\caption{Table shows the accuracy of fine-tuned BERT}\label{tab1}
\begin{center}
\begin{tabular}{|c@{\hspace{5pt}}|c@{\hspace{5pt}}|c@{\hspace{5pt}}|}
\hline
Fine-tune Settings &  Modified ListOps & ListOps\\
\hline
Fine-tuned-layer-12  &  80.8\% & 52.8\% \\
\hline
Fine-tuned-layer-10  &  86.1\% & 54.5\% \\
\hline
Fully-fine-tuned  & 95.1\% & 61.6\% \\
\hline
\end{tabular}
\end{center}
\end{table}

\subsection{Heatmap Analysis} 

To obtain a thorough understanding of the tokens on which the model focuses and pinpoint the most attended tokens in a specific layer, we create heatmaps of the attention scores. These heatmaps offer a visual representation of the model's attention patterns, facilitating the identification of areas with high or low focus.

\begin{figure}[htbp]
    \centering
    \includegraphics[scale=0.3]{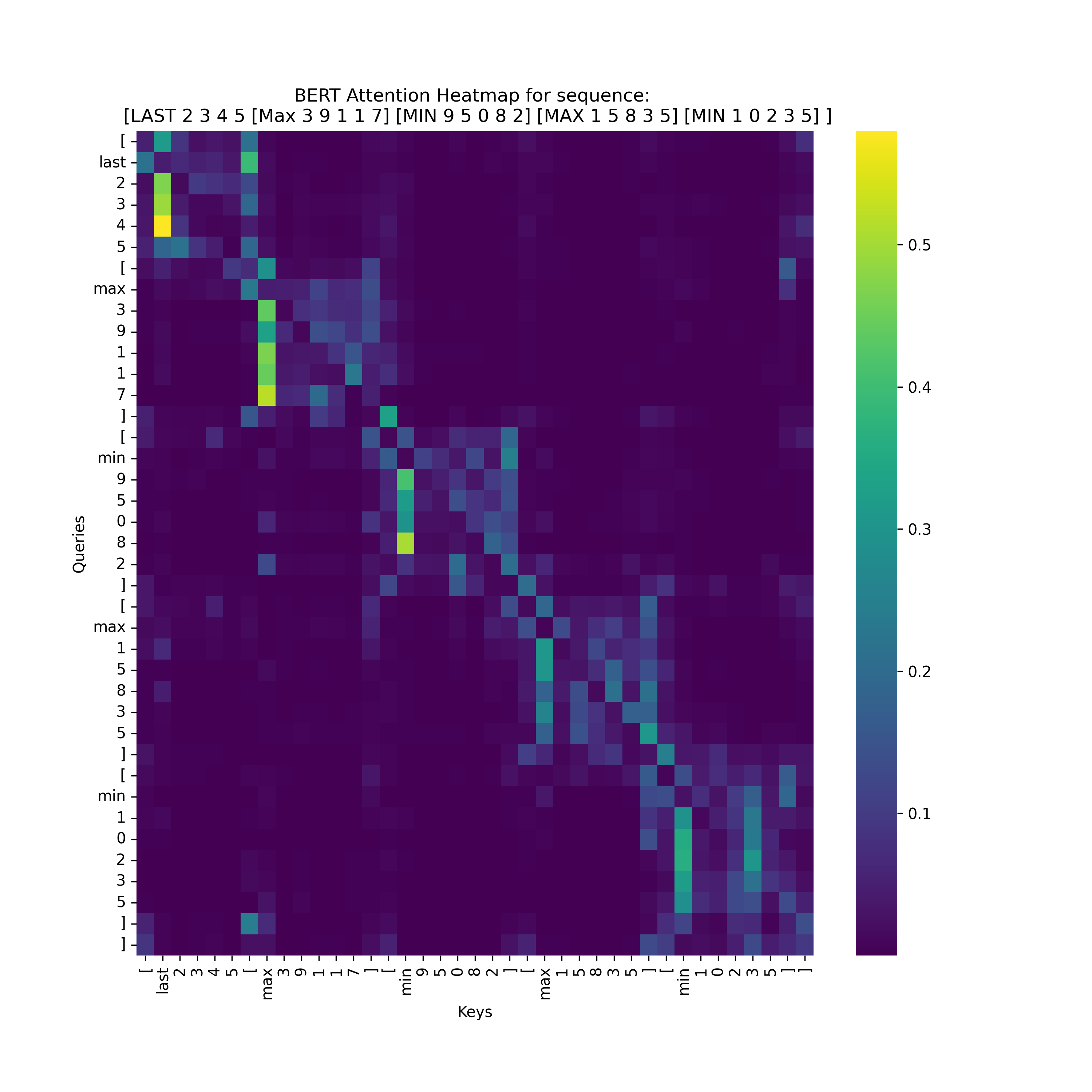}
    \caption{The attention heat map in the 6th layer: the tokens inside each sub-sequence are attending to each other and to their operators, forming attention blocks.}
    \label{fig10}
\end{figure}

\begin{figure}[htbp]
    \centering
    \includegraphics[scale=0.3]{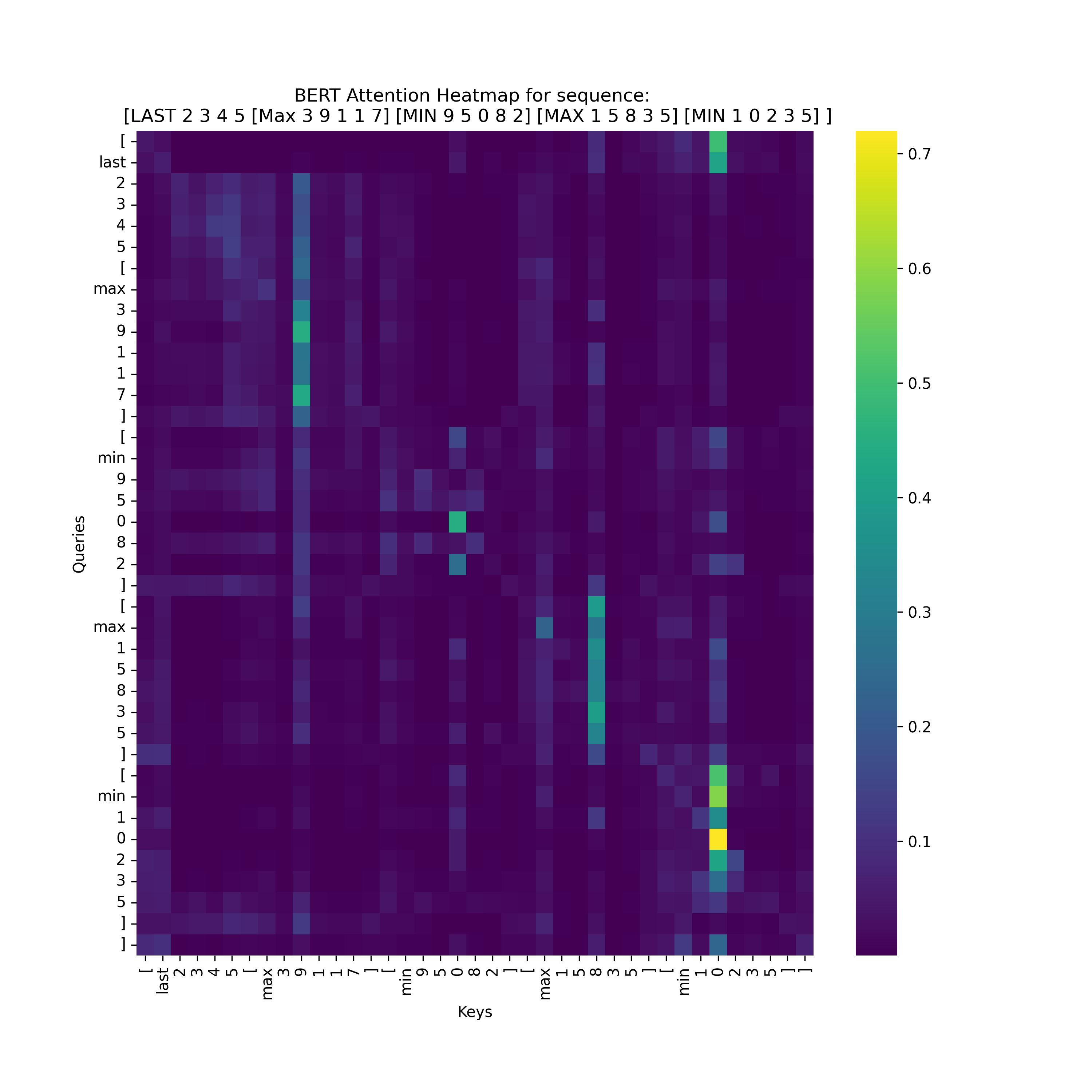}
    \caption{The attention heat map in the 10th layer: the tokens that represent the correct answer in each sub-sequence are the most attended to.}
    \label{fig11}
\end{figure}

\begin{figure}[htbp]
\centering
\includegraphics[scale=0.3]{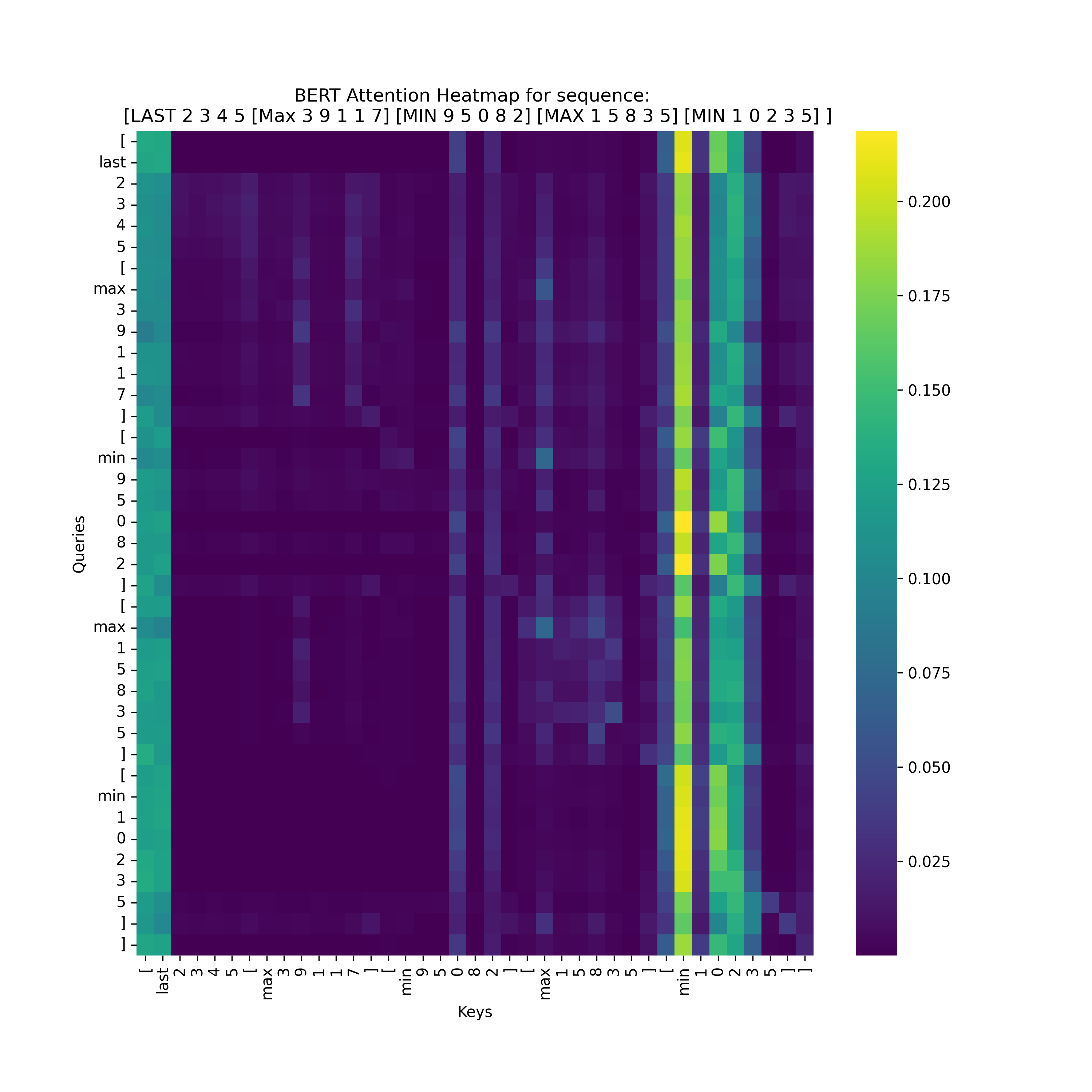}
\caption{The attention heat map in layer 12 (last layer): token ``0", which is the correct answer among other noises are the most attended to tokens.} 
\label{fig12}
\end{figure}

We employed the same formula for computing the attention map as used in the token-to-token analysis discussed earlier. As a result, we expect similar findings but with a distinct visualization method. The heatmaps display the attention of all tokens concurrently in a single image, enhancing the transparency and interpretability of the analysis. This comprehensive perspective enables us to better comprehend the model's attention patterns throughout the entire input sequence. Moreover, the heatmap allows for a clearer visualization of longer sequences, ensuring that the conclusions drawn from previous sections remain valid for extended sequences. The example shown in figure \ref{fig10}, \ref{fig11}, and \ref{fig12} is:

{$$\small{[\text{LAST } \ 2 \ 3 \ 4 \ 5 \  [\text{MAX} \ 3 \ 9 \ 1 \ 1 \ 7 ] \ [\text{MIN} \ 9 \ 5 \ 0 \ 8 \ 2 ] \ [\text{MAX} \ 1 \ 5 \ 8 \ 3 \ 5 \ ] [\text{MIN} \ 1 \ 0 \ 2 \ 3 \ 5 \ ] \ ] }$$}

We will concentrate on layers 6, 10, and 12, where we drew significant conclusions in the previous section. In the generated heatmaps, lighter colors indicate higher attention scores. It is essential to note that the y-axis represents queries, while the x-axis represents keys. Attention heatmaps are not asymmetric: the horizontal axis of a token illustrates what it is attending to, while the vertical axis reveals which tokens it is attended by. For example, if the horizontal axis of the token ``MAX" is entirely yellow, this indicates that it is attending to all tokens. Conversely, if the vertical axis of the same token ``MAX" is yellow, it means that it is being attended to by all tokens.

In layer 6, as we anticipate each token to attend to its respective operator of each sub-sequence, we generate an attention heatmap for a longer sequence containing five operators. Indeed, as expected, figure \ref{fig10} displays five blocks, corresponding to the number of operators, where each block refers to a sub-sequence signifying that the model is dividing the sequences into sub-sequences. A vertical green line is also visible at the beginning of each block, demonstrating that the most attended token is the operator for each sub-sequence.

For layer 10, the results once again align with our previous conclusions, as evidenced by the heatmap in figure \ref{fig11}. In most blocks (sub-sequences), the most attended token is the correct token for the sub-sequence. For example, token 9 of the first ``MAX" operator in the sequence is primarily attended to within its sub-sequence, similar to zero in the first ``MIN" sub-sequence and 8 in the second ``MAX" sub-sequence. Ultimately, the overall most attended token is the correct answer for the entire sequence, which is zero.

Lastly, for the final layer depicted in figure \ref{fig12}, we observe that although the correct token 0 is attended to by most tokens, there are additional tokens that receive attention even though they are not crucial for the answer, such as tokens 2 and 3 in the last sub-sequence. These findings support our earlier conclusion from the token-to-token analysis, where we proposed that some noise exists in the last layer, making fine-tuning with the 10th layer achieving higher results.

\subsection{Entropy Analysis} 

\begin{figure}[htbp]
    \centering
    \includegraphics[scale=0.5]{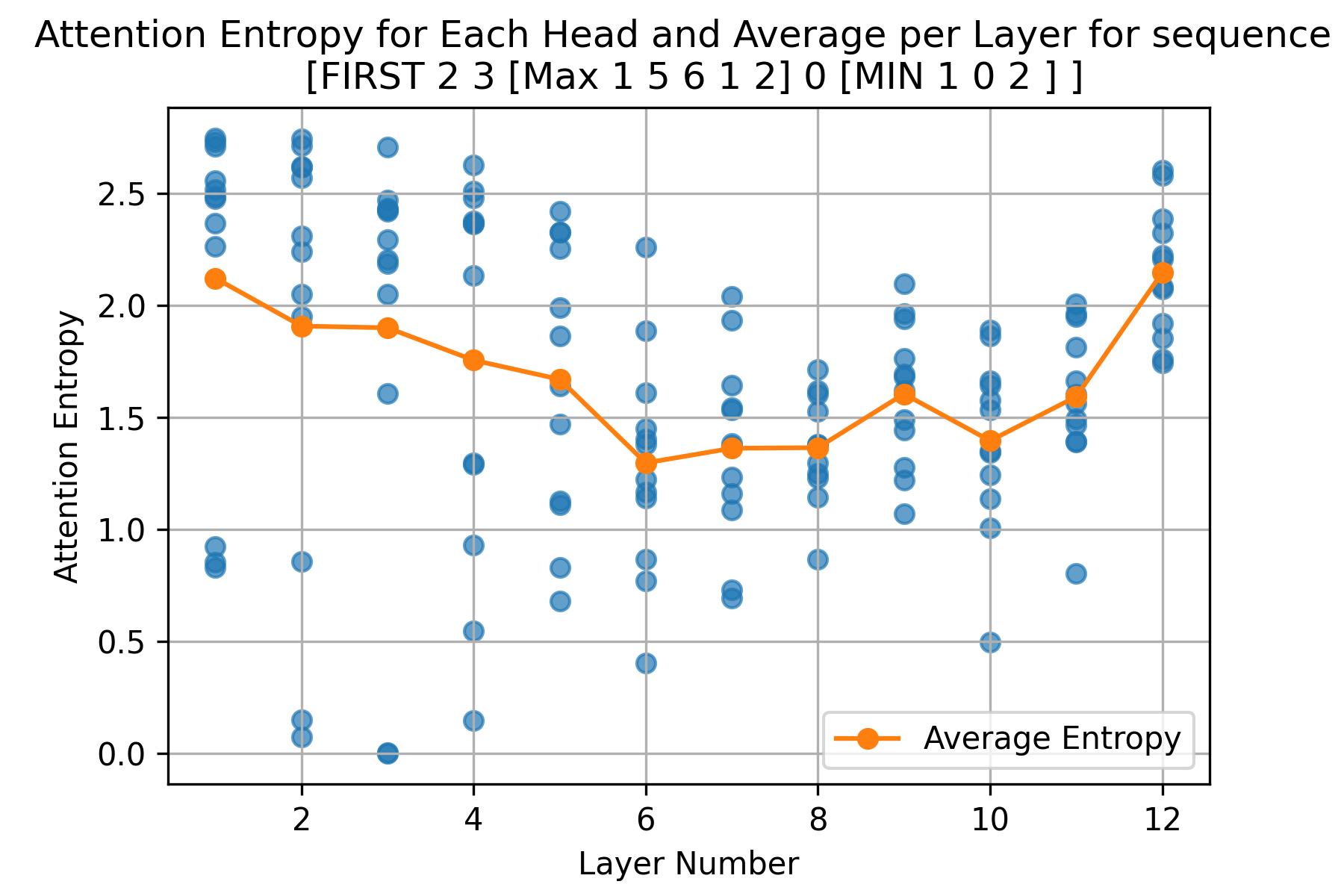}
    \caption{Figure shows the entropy scores of each attention head in blue, as well as their layer-wise average in orange Notably, the lowest entropy values are predominantly found in layers 6 to 11.}
    \label{fig7}
\end{figure}

\begin{figure}[htbp]
    \centering
    \includegraphics[scale=0.5]{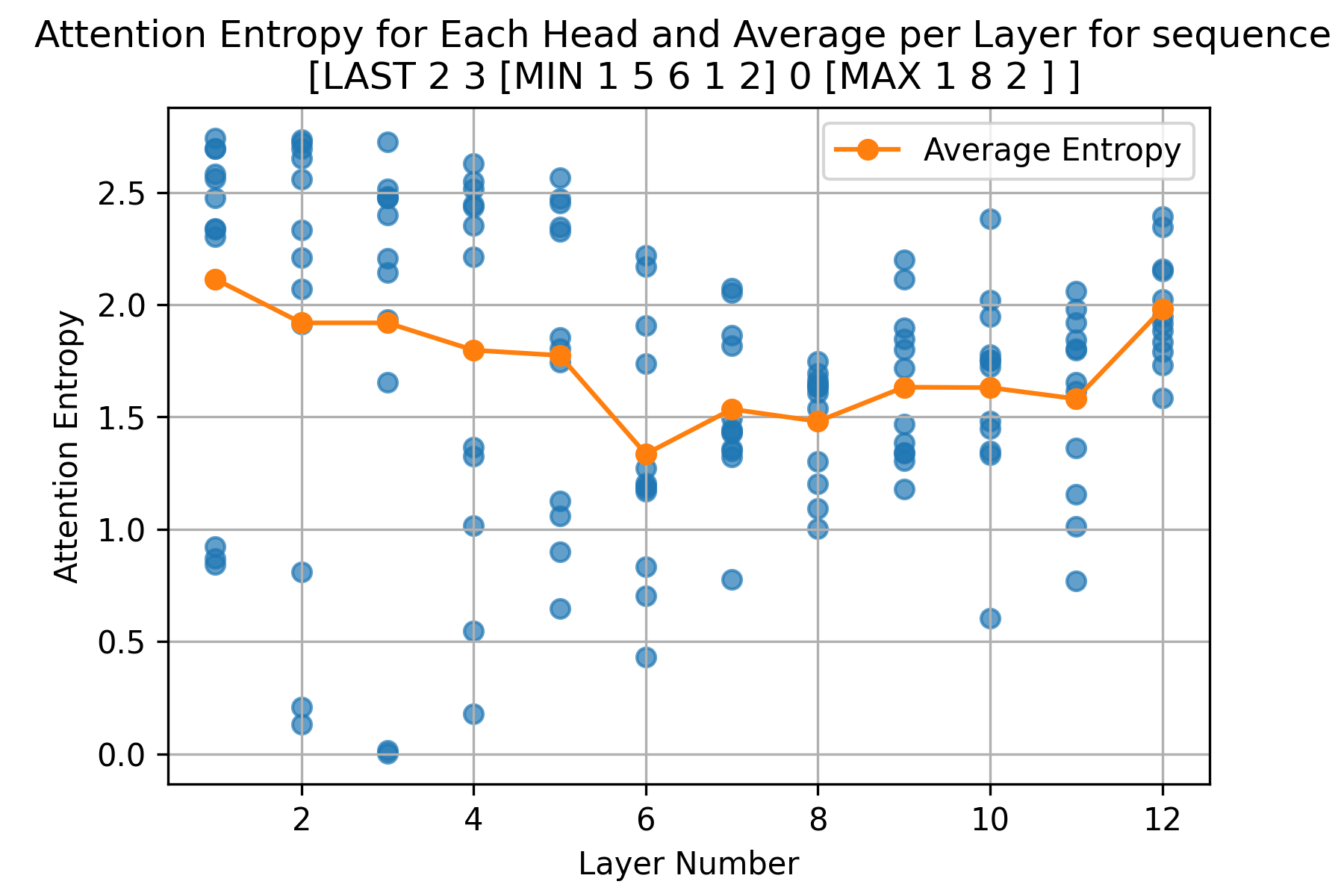}
    \caption{Figure shows the entropy scores of each attention head in blue, as well as their layer-wise average in orange Notably, the lowest entropy values are predominantly found in layers 6 to 11.}
    \label{fig8}
\end{figure}

To solidify our attention analysis, we cross validate it with entropy analysis. By measuring the entropy of attention heads, we can determine whether the attention heads in each layer exhibit focused or broad attention. Theoretically, high entropy means that the attention is spread more evenly across the input elements, implying that the model is uncertain about which element is more important in the given context. This can also mean that the attention head is not specialized, as it does not focus on specific parts of the input. Low entropy, on the other hand, means that the attention is more focused or peaked on certain input elements, indicating that the model is more certain about which elements are important. This can signify that the attention head is specialized, as it concentrates on specific aspects of the input. Entropy in attention is calculated as the sum of attention weights multiplied by their logarithms, reflecting the uncertainty or dispersion of the attention distribution across input elements. Figure \ref{fig7} and \ref{fig8} show the entropy average across 12 layers for the two analyzed sequences in the attention analysis. 

The findings align with what we observed during the attention analysis: layer 6 is evidently specialized in identifying the sub-sequences where all tokens attend to their respective operators. The entropy plot reveals that layer 6 has the lowest entropy for both analyzed sequences. Additionally, layer 10 also exhibits a low entropy score, which can be attributed to its role in selecting the correct answer for each sub-sequence, as indicated in the attention analysis. Lastly, the broad attention in the final layer could be attributed to the model's attempt to attend to all tokens and piece them together in order to predict the correct answer.

\begin{figure}[htbp]
\centering
\includegraphics[scale=0.5]{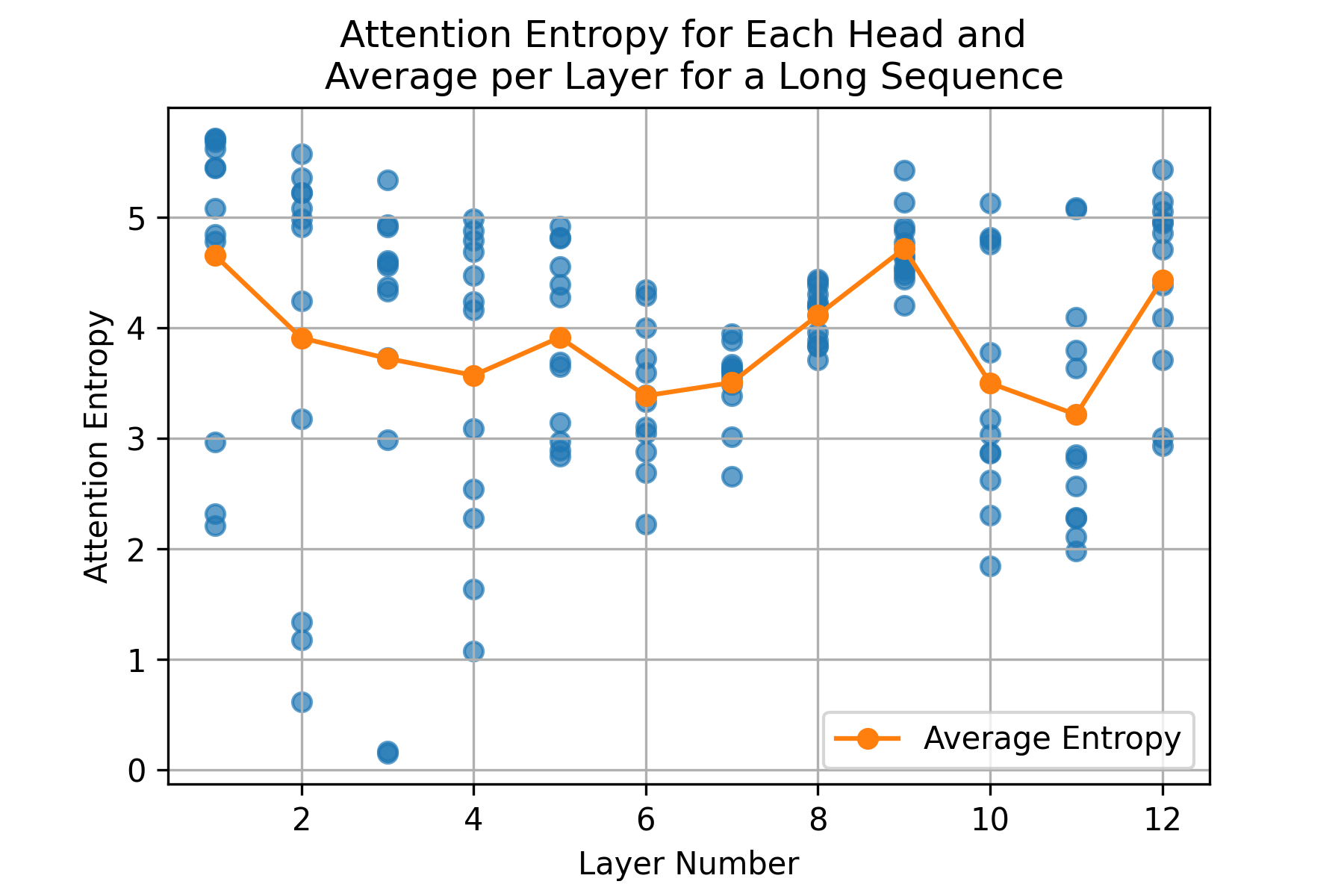}
\caption{ Figure shows the entropy scores of each attention head in blue, as well as their layer-wise average in orange, for a long sequence. Notably, the lowest entropy values are predominantly found in layers 6, 10 and 11.} \label{fig9}
\end{figure}

Moreover, figure \ref{fig9} confirms that our entropy analysis remains consistent not only for the analyzed short sequence but also for a longer sequence from the validation dataset. The same characteristics discussed earlier persist, with layers 6 and 10 displaying some of the lowest entropy scores.  

\subsection{Representation Analysis}

Finally, in this section we analyze the similarity between the embeddings of the tokens to gain more insight about the decision making in BERT for non-language tasks. In a language model, when tokens have similar representations, it means that their corresponding embedding vectors in the model's vector space are close to each other. These similar representations generally indicate that the language model has learned to associate the tokens with similar semantic or syntactic properties, and they may have comparable roles or meanings in the context of the sentences they appear in. Tokens with similar roles, functions, or meanings are likely to have similar representations in the embedding space. We aim to determine whether this principle also holds for non-language tasks. To conduct the similarity analysis, we compute a square similarity matrix. Consider the hidden state matrix H for a particular layer, where each row represents the hidden state for a token in the input sequence. The dot product with its transpose can be written as:

\begin{equation}
S = HH^\top
\end{equation}

In this equation, $S$ is the resulting square similarity matrix, and $H^\top$ represents the transpose of the hidden state matrix $H$. The element at position $(i, j)$ in the matrix $S$ corresponds to the similarity between the $i$-th and $j$-th token representations in the hidden state matrix $H$. Lastly, we utilize a heatmap to visualize this square matrix, illustrating the similarity between tokens. In the heatmaps, dark blue represents a high similarity between tokens, while the color gradient transitioning to light yellow signifies decreasing similarity between them.

\begin{figure}[htbp]
\centering
\includegraphics[scale=0.35]{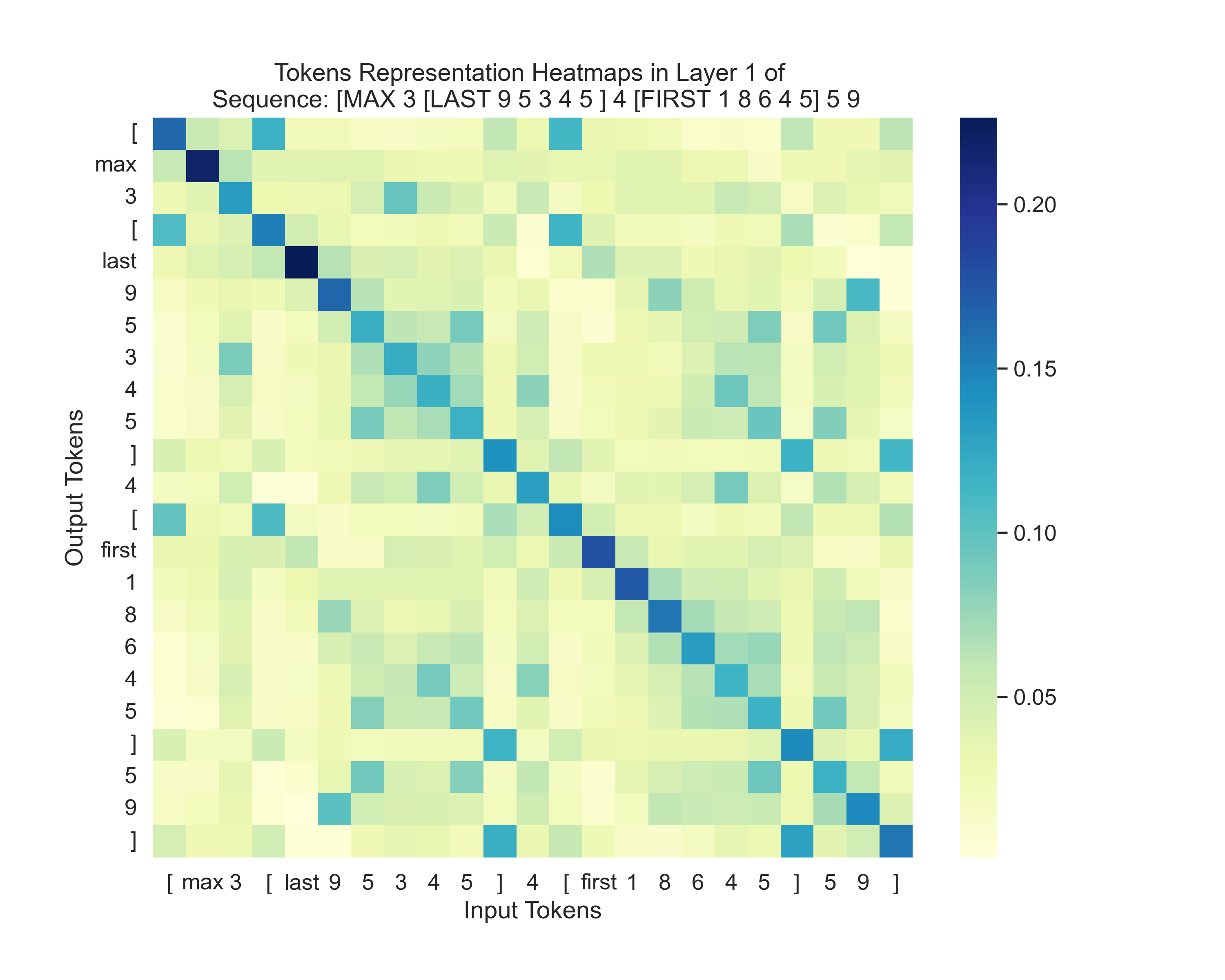}
\caption{Figure illustrates that in the initial layers (layer 1, as depicted in the figure), tokens representing the numbers exhibit similar representations.} 
\label{fig17}
\end{figure}

\begin{figure}[htbp]
    \centering
    \includegraphics[scale=0.35]{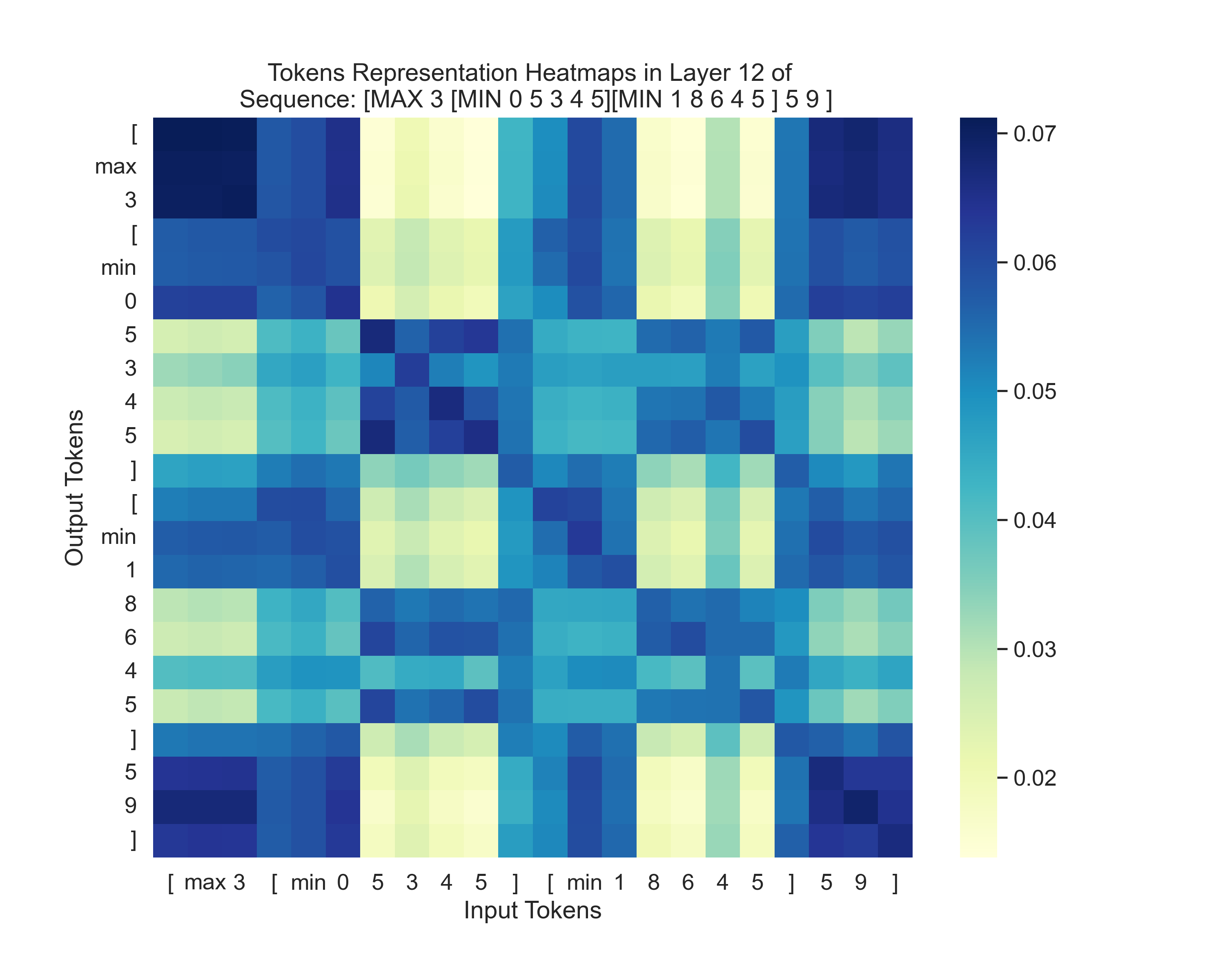}
    \caption{Figure shows that at late layers (in this figure layer 12) the tokens that are similar to each other are the important tokens that are responsible for predicting the correct answer.}
    \label{fig18}
\end{figure}

In the initial layers, there is a strong similarity between numbers across the entire sequence, as depicted in figure \ref{fig17}. This is expected, as the early layers associate numbers with each other, deriving this property from the pre-trained embeddings. However, the intriguing result is that by the final layer, the model has learned to associate the tokens responsible for the prediction together. Referring back to figure \ref{fig1} with step 2, where we show the last step before solving the problem, we can observe that it successfully correlates the essential tokens for the prediction, simplifying the sequence correctly. This analysis demonstrates that the model can solve the task hierarchically, revealing the relationship between these tokens, as illustrated in figure \ref{fig18}. Therefore, similar to how the model associates comparable roles or meanings in language tasks, it connects similar embeddings for tokens most crucial for answer prediction in this task.

The sequence shown in figure \ref{fig18} is: 

$${[\text{MAX } \ 3 \  [\text{MIN} \ 0 \ 5 \ 3 \ 4 \ 5 ] [\text{MIN} \ 1 \ 8 \ 6 \ 4 \ 5 \ ] \ 5 \ 9 \ ] }$$ 

The simplified version after solving all of the sub-sequences with only the last operator to be solved should be:

$${[\text{MAX } \ 3 \ 0 \ 1 \ 5 \ 9 \ ] }$$ 

\begin{figure}[htbp]
    \centering
    \includegraphics[scale=0.35]{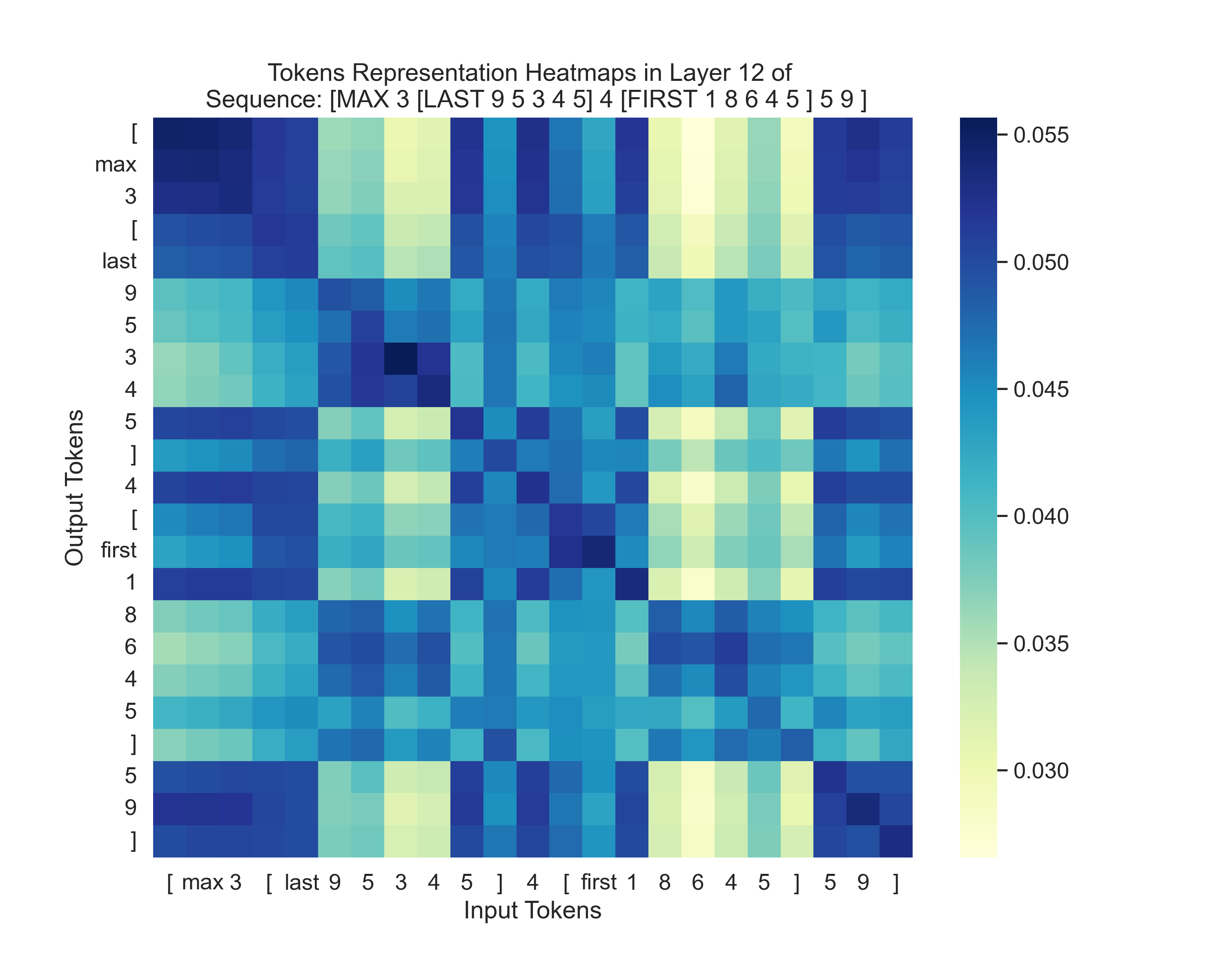}
    \caption{Figure shows that at late layers (in this figure layer 12) the tokens that are similar to each other are the important tokens that are responsible for predicting the correct answer.}
    \label{fig19}
\end{figure}

Upon examining the similarity between these tokens, figure \ref{fig18} evidently shows that they are most closely related to each other (including the sub-sequence operators). As a result, we claim that the model can solve the task in a manner similar to how a human would approach it, using a hierarchical strategy. We provide another example of this behavior in figure \ref{fig19}, where we observe that the input sequence has been simplified from:

$${[\text{MAX } \ 3 \  [\text{LAST} \ 9 \ 5 \ 3 \ 4 \ 5 ] \ 4 \ [\text{FIRST} \ 1 \ 8 \ 6 \ 4 \ 5 \ ] \ 5 \ 9 \ ] }$$ 

to :

$${[\text{MAX } \ 3 \ 5  \ 4 \ 1 \ 5 \ 9 \ ] }$$ 

This is a common pattern that we see along many examples.

\section{Generalization to Long Sequences}

Based on the analyses and conclusions presented above, and considering that the model can capture these types of strategies to solve the ListOps problem, we expect the model, when trained on short sequences, to generalize to longer sequences. To determine whether the model can indeed generalize to longer sequences, we compare two additional settings to the original one, where we trained and tested on sequences of length 200-400. The first setting involves training on sequences of length 20-50 and testing on sequences of length 50-200, while the second entails testing on sequences of length 200-400.

Table \ref{tab2} and \ref{tab3} display the results of the fully fine-tuned BERT on the original ListOps and the modified ListOps, respectively. The findings indeed demonstrate the model's ability to generalize, particularly when trained on sequences of length 20-50 and tested on sequences of length 50-200 for both datasets. For example, the model achieves a 76.1\% score on the modified ListOps dataset when trained on sequences of length 20-50 and tested on sequences of length 50-200, significantly surpassing the 10\% random chance. The model's performance declines when tested on longer sequences, reaching 52.5\%, but this score remains well above the 10\% random chance. In conclusion, due to the strategy employed by the model, as explained in the previous analysis, the model can generalize and achieve satisfactory scores on test datasets that are substantially longer than the sequences it was trained on.  

\begin{table}[htbp]
\caption{ Table shows the generalization results on longer sequences on the original ListOps dataset}
\label{tab2}
\begin{center}
\begin{tabular}{|c@{\hspace{5pt}}|c@{\hspace{5pt}}|c@{\hspace{5pt}}|}
\hline
Trained on sequence length of &  Tested on sequence length of & Accuracy\\
\hline
200-400  &  200-400 & 61.6\% \\
\hline
20-50  &  20-50 & 68.0\% \\
\hline
20-50  &  50-200 & 47.2\%\\
\hline
20-50  & 200-400 & 30.1\% \\
\hline
\end{tabular}
\end{center}
\end{table}

\begin{table}[htbp]
\caption{Table shows the generalization results on longer sequences on the modified ListOps dataset}
\label{tab3}
\begin{center}
\begin{tabular}{|c@{\hspace{5pt}}|c@{\hspace{5pt}}|c@{\hspace{5pt}}|}
\hline
Trained on sequence length of &  Tested on sequence length of & Accuracy\\
\hline
200-400  &  200-400 & 95.1\% \\
\hline
20-50  &  20-50 & 99.9\% \\
\hline
20-50  &  50-200 & 76.1\% \\
\hline
20-50  & 200-400 & 52.5\% \\
\hline
\end{tabular}
\end{center}
\end{table}

\section{Limitation of Attention Analysis}

\begin{figure}
    \centering
    \begin{subfigure}[b]{0.45\textwidth}
        \includegraphics[scale=0.25]{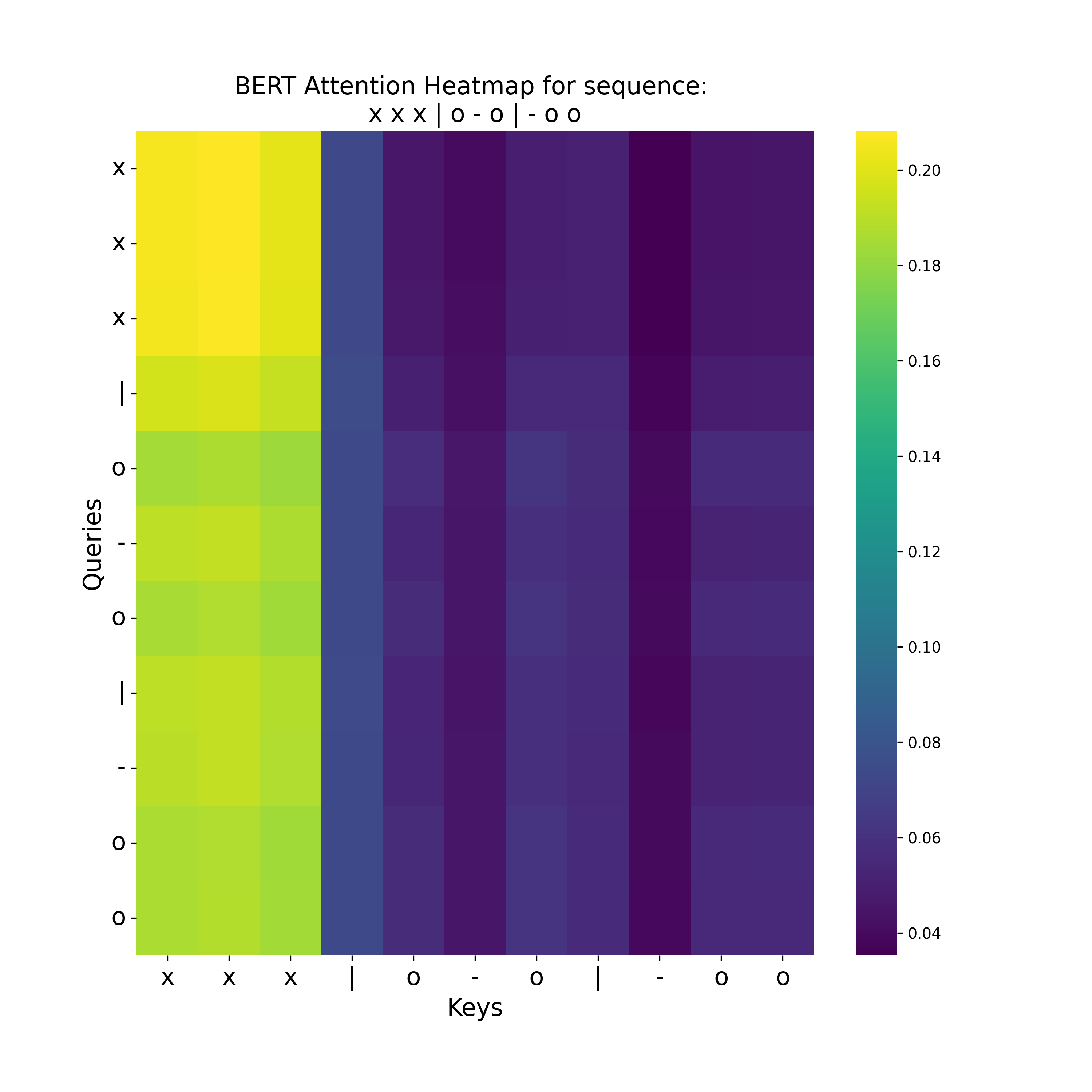}
        \vspace{-0.5cm}
        \label{fig13}
    \end{subfigure}
    \hfill
    \begin{subfigure}[b]{0.45\textwidth}
        \includegraphics[scale=0.25]{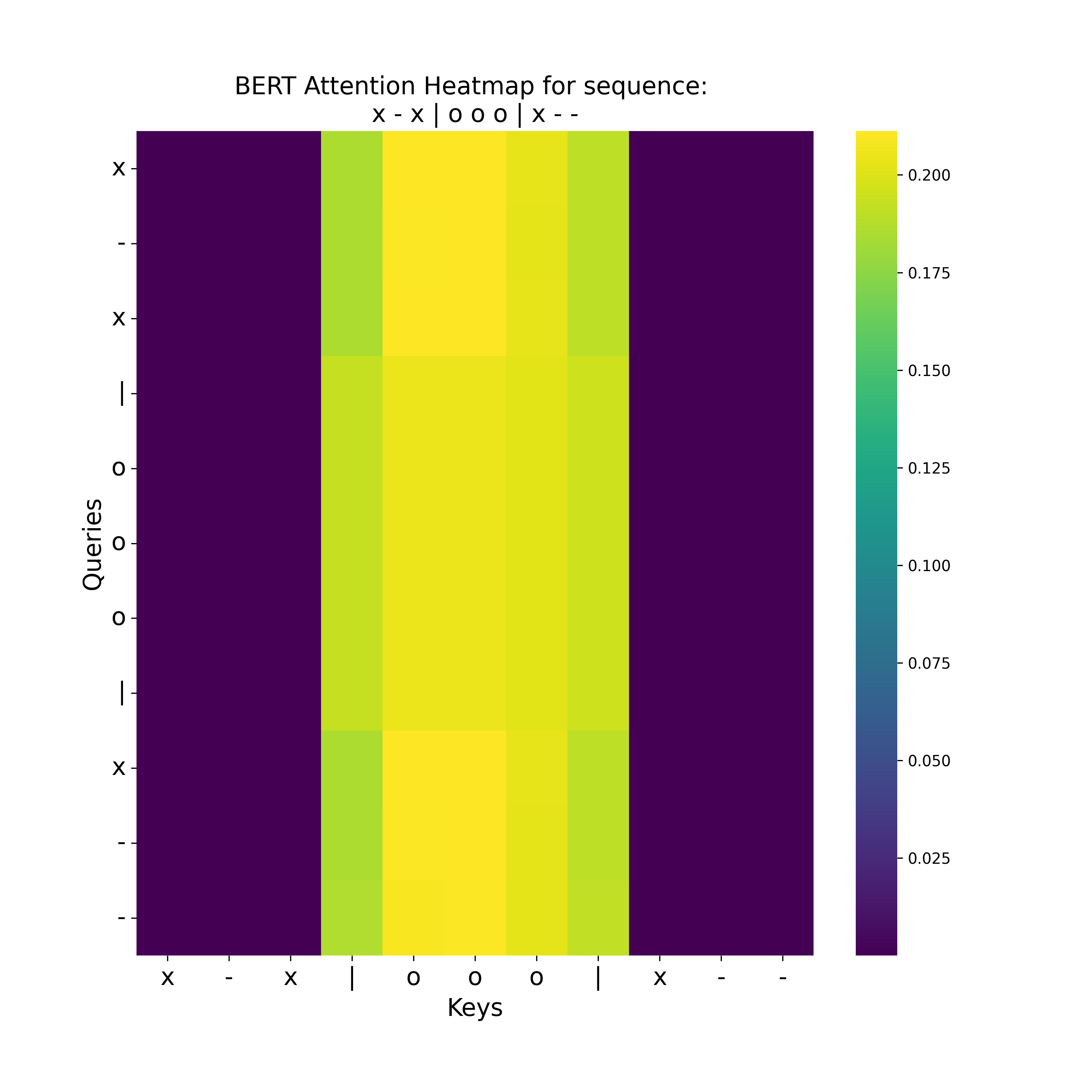}
        \vspace{-0.5cm}
        \label{fig14}
    \end{subfigure}
    \caption{Figure shows the ability of the model to attend to winner when the winner fills horizontal line }
    \label{fig13-14}
\end{figure}

\begin{figure}    
    \begin{subfigure}[b]{0.45\textwidth}
        \includegraphics[scale=0.25]{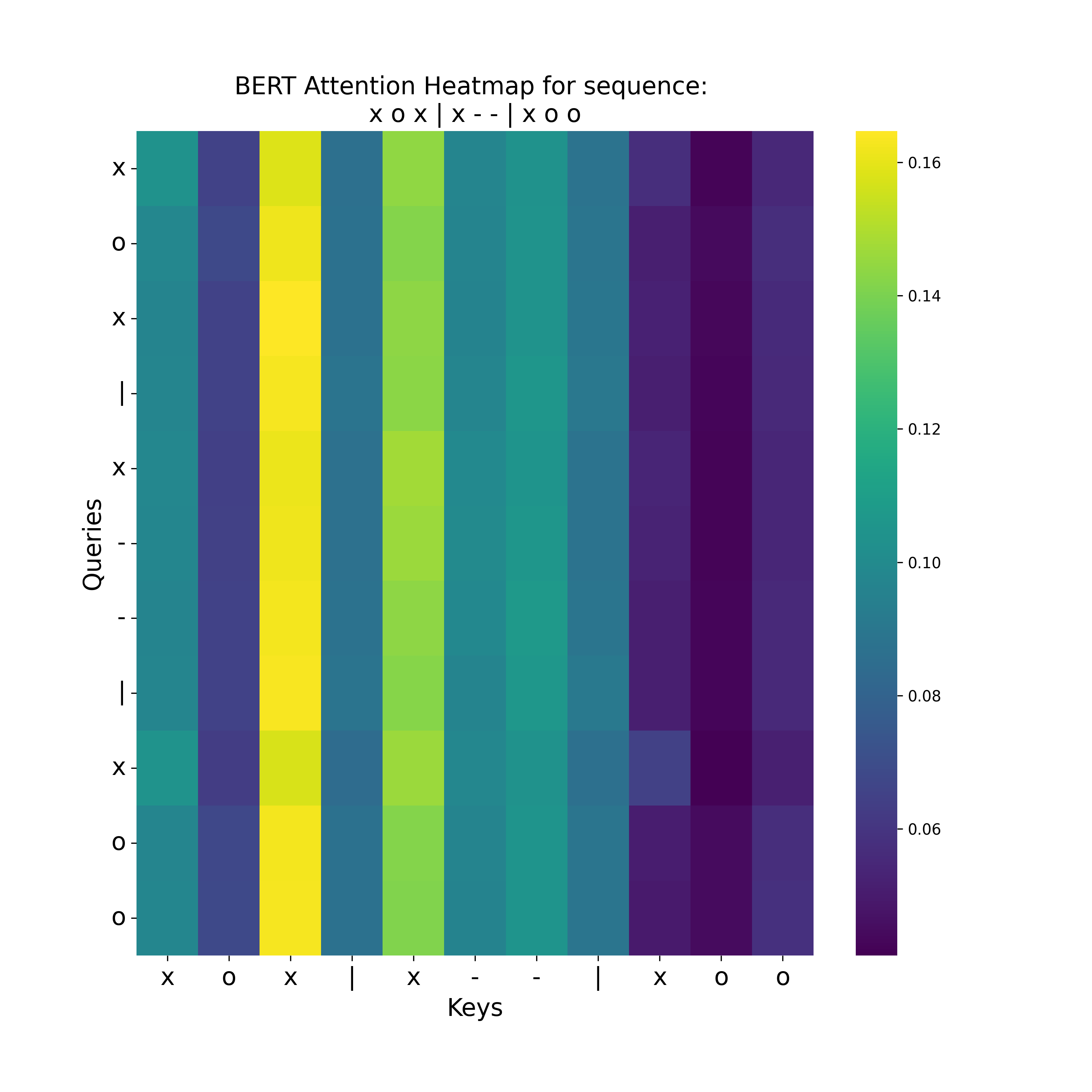}
        \vspace{-0.5cm}
        \label{fig15}
    \end{subfigure}
    \hfill
    \begin{subfigure}[b]{0.45\textwidth}
        \includegraphics[scale=0.25]{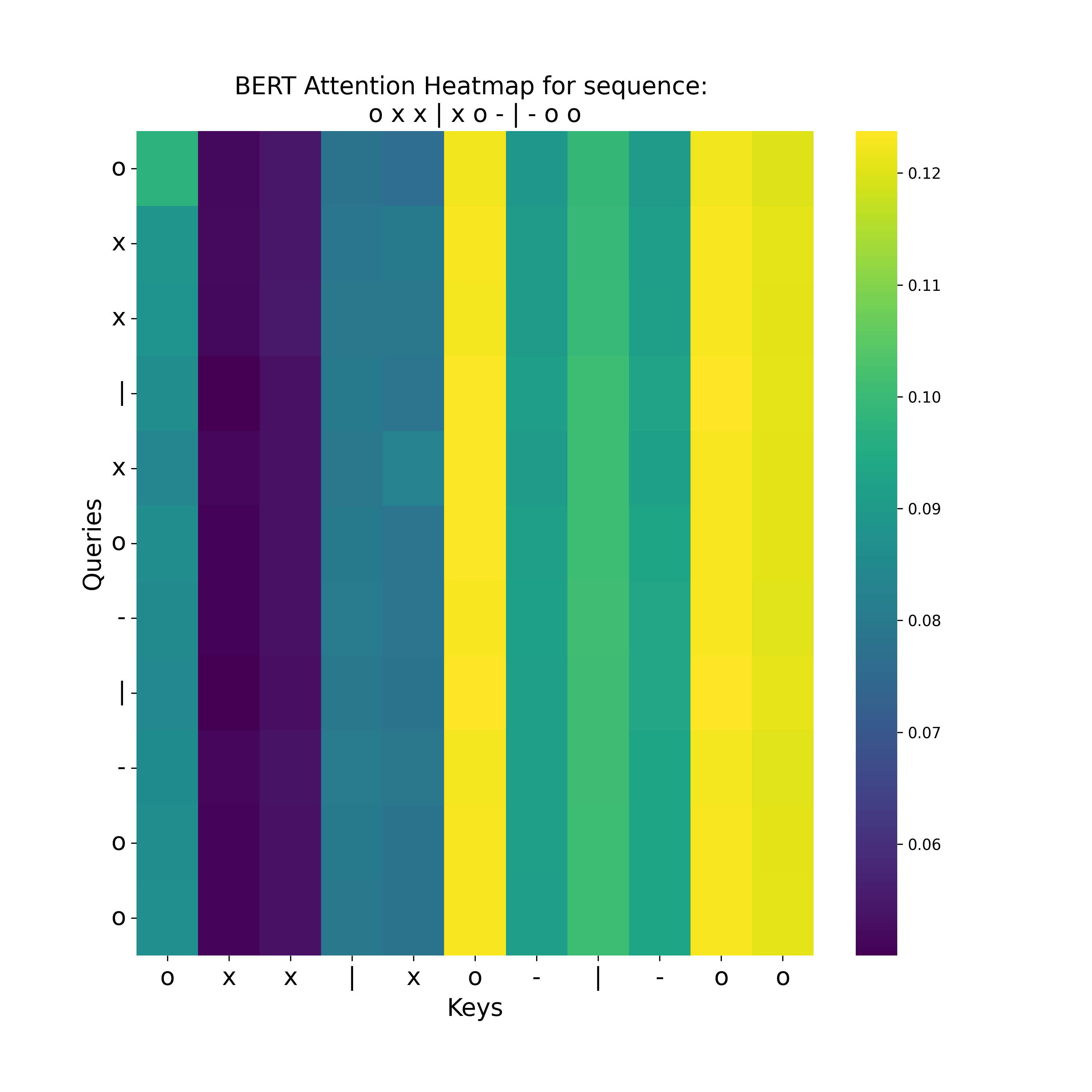}
        \vspace{-0.5cm}
        \label{fig16}
    \end{subfigure}
    \caption{Figure shows that the model cannot detect the tokens responsible for the win, but still attends to the winner}
    \label{fig15-16}
\end{figure}

In this section, we examine the attention of a BERT model fine-tuned to predict the winner of a Tic Tac Toe game. This simple 2D game has a 3x3 dimension, which we flatten into a 1D sequence. The model achieves 100\% accuracy on this game, demonstrating that it can easily predict the winner if a player (``x" or ``o") completely fills a horizontal, vertical, or diagonal line. Through this experiment, we investigate whether the attention could generalize to a 2D input. The model perfectly attends to the correct token when the winner fills a horizontal line, as shown in figure \ref{fig13-14}, but struggles to attend to the correct tokens when the winner fills vertical or diagonal lines, as illustrated in figure \ref{fig15-16}. The difficulty in attending to the correct tokens for vertical and diagonal lines could be due to the way the 2D structure is flattened into a 1D sequence, making it more challenging for the model to recognize vertical and diagonal patterns, even in short sequences. This limitation is expected for a model like BERT, as it was trained on 1D language data and may not generalize well to 2D data structures. However, even though the model does not attend to the correct tokens that determine the game's winner when the lines are diagonal or vertical, the heatmaps reveal that it mostly attends to some tokens associated with the actual winner. This could suggest that the attention is present but not clearly depicted in the illustrations.   

\section{Conclusion and Outlook}

In conclusion, this paper presents novel probing tasks to comprehend the steps that BERT model follows in solving non-language tasks. In contrast to vision and language tasks, where we lack an understanding of human processing, the carefully chosen tasks involve a hierarchical approach with straightforward steps. Our analysis includes token-to-token examination, through which we discover that specific layers have designated tasks. For example, layer 6 identifies sub-sequences within the entire sequence, while layer 10 simplifies the input sequence into a more straightforward sequence, similar to how humans might approach the problem. These observations are supported by entropy calculations for each layer, with layers 6 and 10 exhibiting lower scores, indicating their specialized tasks. These observations have led to the identification of an unconventional yet efficient fine-tuning approach, in which we only fine-tune the parameters of layer 10 of the model, resulting in comparatively good performance. Additionally, our representation analysis demonstrates that the model can assign similar embeddings to tokens crucial for accurately answering the sequence. Lastly, we highlight a limitation of the analysis, as attention mechanisms only partially succeed in identifying 2D patterns.

This paper could serve as a base for future work in this area, particularly concerning analyzing attention for non-language tasks, as these models are increasingly being employed for tasks beyond language processing. Additionally, we hope that this study inspires the development of novel fine-tuning techniques. We believe that this work could be expanded by examining more non-language tasks, where insights gained from the attention outputs could contribute to a deeper understanding of the underlying mechanisms in these models. Finally, conducting a psychological study to compare the methods various individuals use to solve the ListOps tasks with the demonstrated steps of the model's problem-solving approach would be interesting. 

\section*{Acknowledgements}

This work was funded by the Deutsche Forschungsgemeinschaft (DFG, German Research Foundation). The cluster used to train the models was also funded by the German Research Foundation (DFG) - 456666331.


%
%
%
%

\end{document}